\documentclass[10pt,journal,compsoc]{IEEEtran}
\ifCLASSOPTIONcompsoc
\else
\fi
\usepackage[utf8]{inputenc} 
\usepackage[T1]{fontenc}    
\usepackage{microtype}
\usepackage{wrapfig}
\usepackage{graphicx}
\usepackage{enumitem}
\usepackage{booktabs} 
\usepackage{url}            
\usepackage{booktabs}       
\usepackage{amsfonts}       
\usepackage{nicefrac}       
\usepackage{microtype}      
\usepackage{amsmath}
\usepackage{amssymb}
\usepackage{caption}
\usepackage{enumitem}
\newcommand{\ie}{i.e.\xspace}
\newcommand{\eg}{e.g.\xspace}
\newcommand{\etal}{et al.\xspace}
\usepackage{xspace}
\usepackage[pagebackref=true,breaklinks=true,letterpaper=true,bookmarks=false]{hyperref} 

\usepackage{bm}
\usepackage{caption}
\usepackage[american]{babel}
\usepackage{amssymb}
\usepackage{amsmath,amsfonts}
\usepackage{amsopn}
\usepackage{bm} 
\usepackage{multirow}

\newcommand{\vct}[1]{\boldsymbol{#1}} 
\newcommand{\mat}[1]{\boldsymbol{#1}} 

\newcommand{\ProbOpr}[1]{\mathbb{#1}}

\newcommand{\expect}[2]{%
\ifthenelse{\equal{#2}{}}{\ProbOpr{E}_{#1}}
{\ifthenelse{\equal{#1}{}}{\ProbOpr{E}\left[#2\right]}{\ProbOpr{E}_{#1}\left[#2\right]}}} 
\newcommand{\var}[2]{%
\ifthenelse{\equal{#2}{}}{\ProbOpr{VAR}_{#1}}
{\ifthenelse{\equal{#1}{}}{\ProbOpr{VAR}\left[#2\right]}{\ProbOpr{VAR}_{#1}\left[#2\right]}}} 

\DeclareMathOperator{\argmax}{arg\,max}
\DeclareMathOperator{\argmin}{arg\,min}

\newcommand{\vtheta}{\vct{\theta}}

\newcommand{\vx}{{\vct{x}}}

\newcommand{\vw}{\vct{w}}

\newcommand{\mW}{\mat{W}}

\newcommand{\eat}[1]{}

\ifCLASSINFOpdf
\else
\fi
\begin{document}
%
\title{Identifying and Compensating for Feature Deviation in Imbalanced Deep Learning}

%
%
%
%

\author{Han-Jia Ye,
	Hong-You Chen,
	De-Chuan Zhan,
	and~Wei-Lun Chao
	\IEEEcompsocitemizethanks{
		\IEEEcompsocthanksitem H.-J. Ye and D.-C. Zhan are with State Key Laboratory for Novel Software Technology, Nanjing University,  Nanjing, 210023, China.
		\protect\\
		Hong-You Chen and Wei-Lun Chao are with Department of Computer Science and Engineering, the Ohio State University, USA.
		\protect\\
		E-mail: yehj@lamda.nju.edu.cn, chen.9301@osu.edu, zhandc@lamda.nju.edu.cn, chao.209@osu.edu}
}

%
%

\markboth{Journal of \LaTeX\ Class Files,~Vol.~Xx, No.~X, Xxxx~20Xx}%
{Ye \MakeLowercase{\textit{et al.}}: Heterogeneous Few-Shot Model Rectification with Semantic Mapping}
%




\IEEEtitleabstractindextext{%
\begin{abstract}
Classifiers trained with class-imbalanced data are known to perform poorly on test data of the ``minor'' classes, of which we have insufficient training data.
In this paper, we investigate learning a ConvNet classifier under such a scenario. We found that a ConvNet significantly {over-fits} the minor classes, which is quite \emph{opposite} to traditional machine learning algorithms that often {under-fit} minor classes. We conducted a series of analysis and discovered the feature deviation phenomenon --- the learned ConvNet generates deviated features between the training and test data of minor classes --- which explains how over-fitting happens. To compensate for the effect of feature deviation which pushes test data toward low decision value regions,  
we propose to incorporate class-dependent temperatures (CDT) in training a ConvNet. CDT \emph{simulates} feature deviation in the training phase, forcing the ConvNet to enlarge the decision values for minor-class data so that it can overcome \emph{real} feature deviation in the test phase. We validate our approach on benchmark datasets and achieve promising performance. We hope that our insights can inspire new ways of thinking in resolving class-imbalanced deep learning.
\end{abstract}

}

\maketitle


%
\IEEEpeerreviewmaketitle

\section{Introduction}
\label{sec:intro}

Convolutional neural networks (ConvNets) have led to remarkable breakthroughs in visual recognition
\cite{he2016deep,huang2017densely,krizhevsky2012imagenet,simonyan2015very,szegedy2015going}, thanks to the large amount of labeled instances available for each object class of interest \cite{deng2009imagenet,russakovsky2015imagenet}. In practice, however, we frequently encounter training data with \emph{imbalanced class distributions}. For example, real-world datasets often have the so-called long-tailed distribution: a few ``{major}'' classes claim most of the instances, while most of the other ``{minor}'' classes are represented by relatively fewer instances~\cite{guo2016ms,krishna2017visual,lin2014microsoft,thomee2015yfcc100m,van2018inaturalist}. ConvNet classifiers trained with this kind of datasets using empirical risk minimization (\eg, stochastic gradient descent on uniformly sampled instances)
have been found to perform poorly on minor classes~\cite{buda2018systematic,liu2019large,van2017devil}.

\begin{figure}[t]
	\centering
	\begin{minipage}[h]{0.7\linewidth}
		\centering \includegraphics[width=\linewidth]{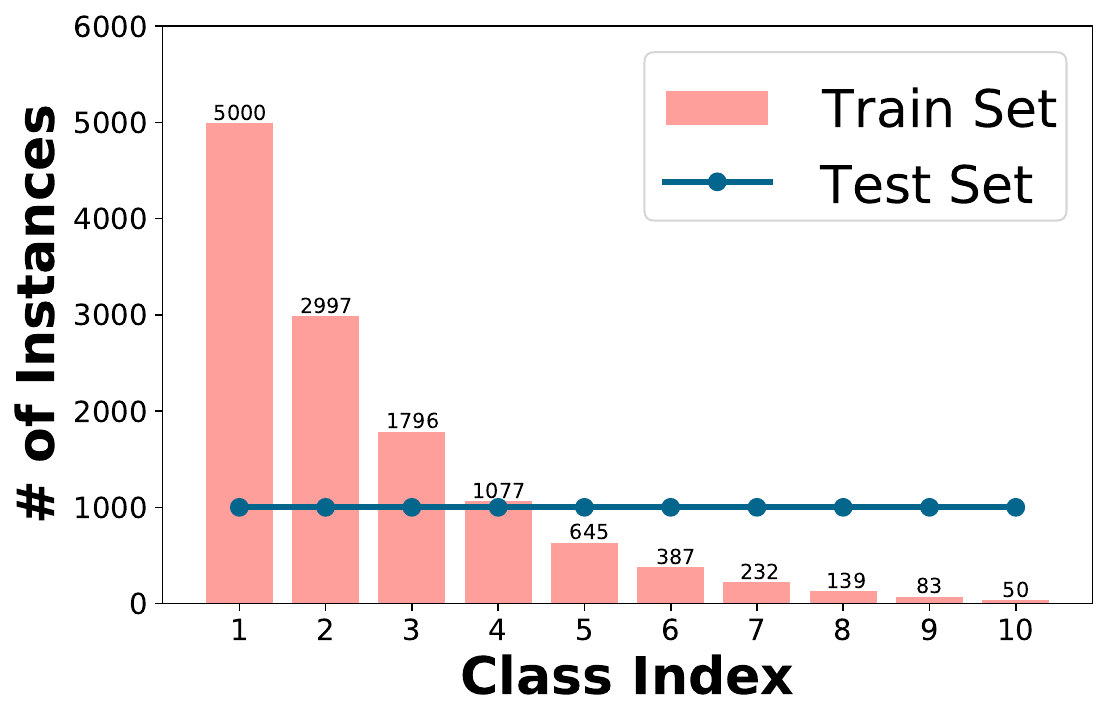}
		\vskip-5pt
		\mbox{\small (a) The number of training and test data per class}
		\\
	\end{minipage}\\
	\vskip 5pt
	\begin{minipage}[h]{0.7\linewidth}
		\centering
		\centering
		\includegraphics[width=\linewidth]{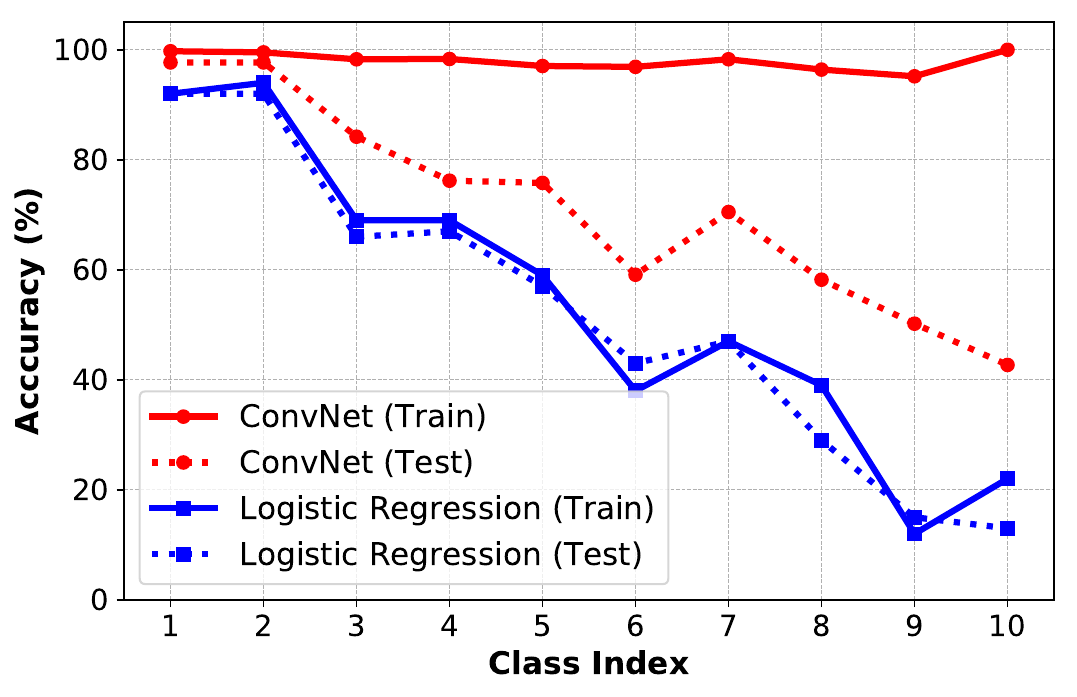}
		\vskip-5pt
		\mbox{\small (b) Training and test accuracy per class}
		\\
	\end{minipage}
\caption{\textbf{Learning with class-imbalanced data:} \textbf{(a)} the number of training (red) and test (blue) instances per class in a class-imbalanced CIFAR-10 dataset \cite{cui2019class,krizhevsky2009learning}; \textbf{(b)} the training (solid) and test (dashed) accuracy per class using a ResNet classifier~\cite{he2016deep} learned end-to-end (red) and a logistic regression learned on pre-defined features (blue). While both classifiers perform poorly on minor classes, the reasons are different. The ResNet suffers over-fitting to minor classes, while logistic regression suffers under-fitting to minor classes.}\label{fig:train_test_line_plot}
\vskip -5pt
\end{figure}

\begin{figure}[t]
\centering
\includegraphics[width=0.9\linewidth]{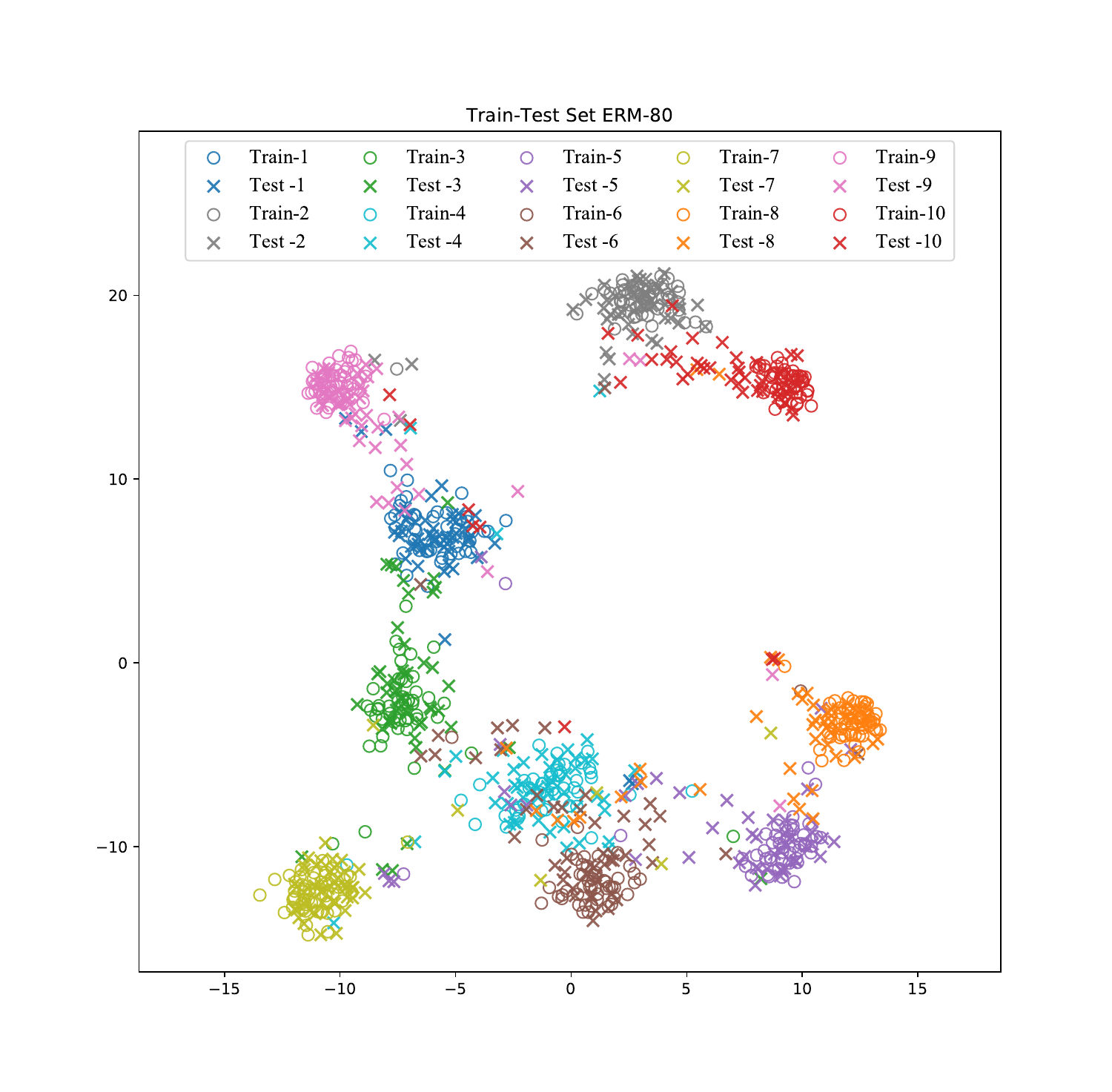}
\caption{\textbf{Feature deviation between training and test data.} Building on the same experiment as in \autoref{fig:train_test_line_plot}, we sample \emph{equal numbers} of training and test data for all classes, and plot the t-SNE embeddings~\cite{maaten2009visualizing} of the training (circle) and test (cross) features before the last fully-connected layer of a ResNet. 
Instances of different classes are colored differently. The smaller the class index is, the more training data it has. There is a notable feature deviation for minor (\eg, red) classes. {We note that since t-SNE derives data-dependent embeddings, it is crucial to sample balanced examples for it; otherwise, major class data will occupy the whole space.}}\label{fig1:feature_d}
\end{figure}

One common explanation of this issue is the dominance of the major class data in training~\cite{he2013imbalanced}. 
Concretely, a classifier trained to minimize the average per-instance loss
tends to bias itself toward predicting the major classes \cite{chawla2002smote,he2009learning,khan2017cost,johnson2019survey,wang2016training}. As a result, the learned classifier would classify most of the test and even the training data of minor classes incorrectly.
To resolve this problem, a variety of approaches have been proposed to {scale up the influence of minor-class data} during training; \eg, by re-sampling or re-weighting \cite{buda2018systematic,byrd2019what,chawla2002smote,ling1998data,shen2016relay,cui2019class,huang2016learning,mahajan2018exploring,wang2016training}. It is worth mentioning that both re-sampling and re-weighting have been widely used before the deep learning era~\cite{he2013imbalanced,he2009learning}.

In this paper, we investigate another explanation
--- \emph{a ConvNet classifier trained with class-imbalanced data {over-fits} the minor class data.} 
This is evidenced by the huge gap between the training and test accuracy of minor-class data in \autoref{fig:train_test_line_plot}\footnote{This was mentioned in~\cite{buda2018systematic,cao2019learning,cui2019class,Kim2020M2M}, when the re-weighting or re-sampling factors were not set properly. Here, we however found that even without re-weighting or re-sampling, over-fitting already occurs.}. Specifically, a ConvNet can easily fit the limited amount of minor class data to attain $\sim 100\%$ training accuracy but just cannot generalize well to the test data. We argue that, \emph{this explanation is fundamentally different from the aforementioned one.} Applying re-weighting or re-sampling to ConvNets thus may not be effective~\cite{kang2019decoupling} and may even \emph{aggravate} the problem of over-fitting to minor classes.

At the first glance, it is not surprising that a ConvNet can fit the training data (almost) perfectly \cite{zhang2016understanding}. What intrigue us are the drastically different behaviors of a \emph{single} ConvNet on test data of different classes. We note that, the problem of over-fitting has been widely studied in machine learning, but mostly from a holistic perspective (\ie, over all classes) or on scarce but class-balanced training data.

To better understand the ``over-fitting to minor classes'' issue, we decompose a ConvNet classifier by
\begin{align}
\hat{y} = \argmax_{c\in\{1,\cdots,C\}} \vw_c^\top f_{\vtheta} (\vx), \label{eq:test}
\end{align}
where $\vx$ is the input, $f_{\vtheta}(\cdot)$ is the feature extractor parameterized by $\vtheta$ (\eg, a ConvNet), and $\vw_c$ is the linear classifier of class $c$.
We observe an interesting phenomenon --- \emph{for a minor class, the features $f_{\vtheta}(\vx)$ of the training and test instances are deviated from each other}, as illustrated in \autoref{fig1:feature_d}\footnote{We emphasize that this deviation is fundamentally different from the domain shift studied in domain adaptation (DA)~\cite{ben2010theory,gong2012geodesic,ganin2016domain}. In DA, the training (source) and test (target) data come from different domains (\eg, synthetic and real images). Here, our training and test data are from the same domain (\eg, CIFAR images). The feature deviation is caused by training ConvNets end-to-end with class-imbalanced data.}. 
The fewer the training instances of a class are, the larger the deviation is. The linear classifiers $\{\vw_c\}_{c=1}^C$ learned to separate the training instances of different classes, therefore, are hardly applicable to separate the test instances of minor classes. 

\begin{figure}[t]
	\centering
	\includegraphics[width=1.0\linewidth]{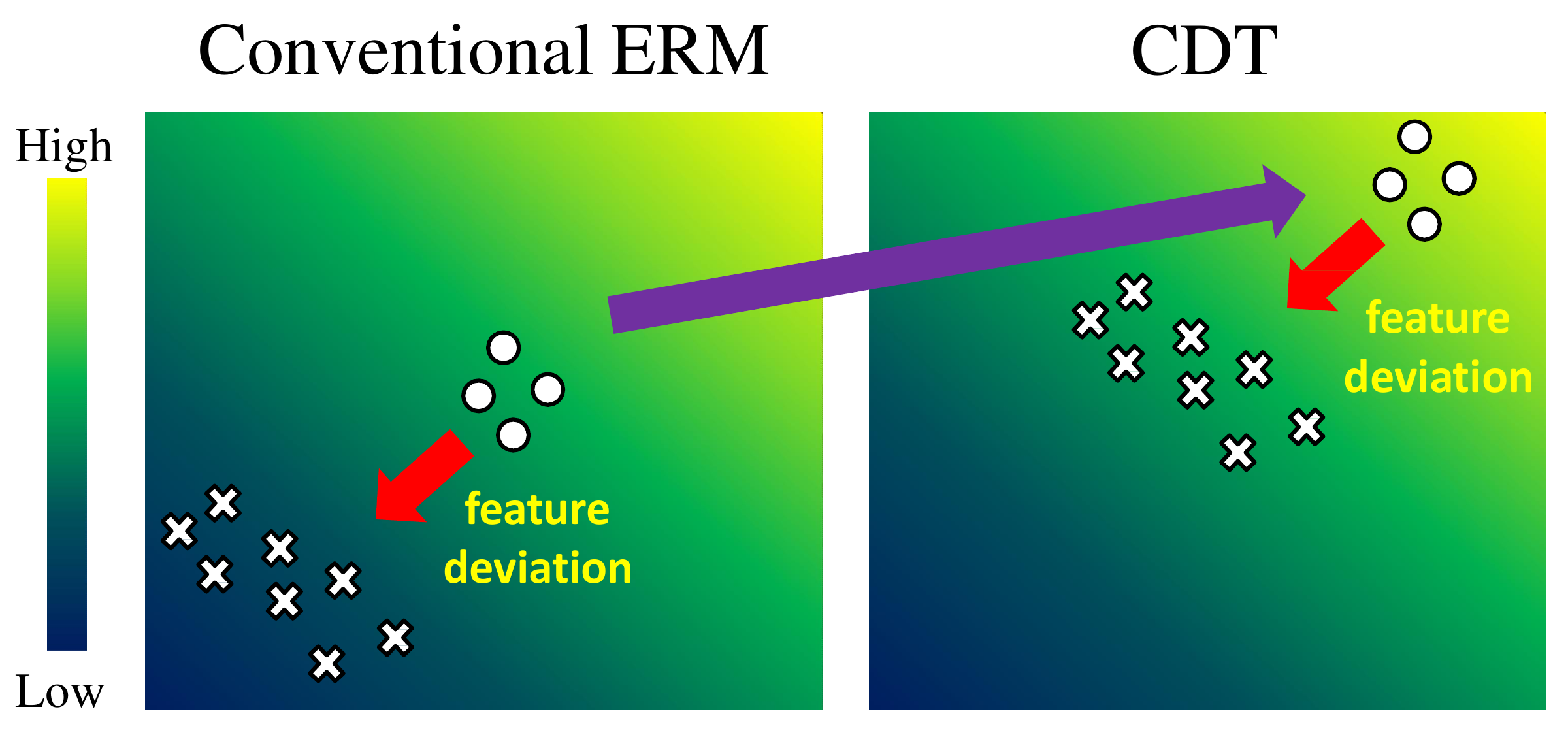}
	\caption{\textbf{The effect of feature deviation and how class-dependent temperatures (CDT) compensate for the effect.}  We show the training/test features (circle/cross) and the decision values $\vw_c^\top f_{\vtheta} (\vx)$ (the background color) of a \emph{minor} class $c$. The red arrows indicate feature deviation: the test features move away from the training features and toward regions of low decision values. CDT compensates for this effect by forcing training data to have higher values (purple arrow). Thus, even after feature deviation, the decision values of the test data are higher than the counterparts in learning with conventional empirical risk minimization (ERM).}
	\label{fig:1:CDT}
\end{figure}

We posit that such a \emph{feature deviation} phenomenon explains the over-fitting problem\footnote{We note that, our preliminary technical report \cite{ye2020identifying} has been on arXiv.org since January, 2020, and we are among the first to observe and systematically study the feature deviation phenomenon.} and accounts for the poor accuracy of a ConvNet classifier on minor class data, and naively scaling up the influence of minor classes in training may worsen the problem. We note that, feature deviation is not commonly seen in traditional machine learning that uses pre-defined features. Thus, traditional algorithms are likely suffer a different problem in class-imbalanced learning, \eg, under-fitting to minor classes (\autoref{fig:train_test_line_plot} (b)), for which scaling up the influence of minor classes can be effective.

We study further how feature deviation affects a ConvNet classifier in making predictions.
We find that when feature deviation occurs (\eg, for class $c$), the test instances of class $c$ tend to move toward regions of lower decision values (\ie, $\vw_c^\top f_{\vtheta} (\vx)$) compared to the training instances, making them hard to be classified correctly. See \autoref{fig:1:CDT} for an illustration. 

To {compensate for this effect}, we propose to incorporate \textbf{class-dependent temperatures (CDT)} in training a ConvNet. The core idea is to \emph{simulate} the effect of feature deviation in training --- intentionally decreasing the decision values of the minor class data --- to encourage the ConvNet classifier to learn to tackle it.
Concretely, let $a_c$ be the temperature of class $c$, which is set inversely proportionally to its number of training data (\ie, minor classes have larger temperatures), we divide the decision value $\vw_c^\top f_{\vtheta} (\vx)$ by $a_c$ in {training}. This in turn forces the ConvNet to increase $\vw_c^\top f_{\vtheta} (\vx)$ by a factor of $a_c$ in minimizing the training error. As a result, even if feature deviation occurs in testing, the test instances will have higher decision values than before (\ie, than learning without CDT), more likely leading to correct predictions.
On several benchmark datasets~\cite{krizhevsky2009learning,le2015tiny,van2018inaturalist}, our approach achieves superior performance compared to the existing methods.

The contributions of this paper are four-folded. 
\begin{itemize}
    \item We conduct extensive analysis on class-imbalanced learning. Specifically, we analyze the ConvNet's over-fitting behavior to minor classes
    and identify the existence of feature deviation between the training and test instances.
    \item We systematically investigate the reasons behind the poor performance of ConvNets and traditional machine learning methods on minor classes, and argue that the widely-applied re-weighting and re-sampling are not effective for ConvNets. 
    \item We propose an effective approach to \emph{compensate for the effect of feature deviation} by incorporating class-dependent temperatures (CDT) in training. {We emphasize that CDT do not reduce feature deviation, but encourage the ConvNet classifier to learn to deal with it.}
    \item We conduct extensive experiments on multiple benchmark datasets. CDT outperforms or is on a part with state-of-the-art approaches in most of the cases.
\end{itemize}

\section{Related Work}
\label{sec:related}

We review approaches on learning with class-imbalanced data. There are two mainstream approaches: \emph{re-sampling} based and \emph{cost-sensitive} based. Please see also \cite{he2013imbalanced,krawczyk2016learning,japkowicz2000class,japkowicz2002the,johnson2019survey,buda2018systematic,he2009learning} for surveys or comprehensive studies.

\noindent\textbf{Re-sampling based methods} change the training data distribution to match the test data\footnote{In testing, one usually assumes \emph{class-balanced} test data or computes average per-class accuracy.}~\cite{drummond2003c4,buda2018systematic,van2017devil}. Popular ways are to over-sample the minor-class instances~\cite{buda2018systematic,byrd2019what,shen2016relay} or under-sample the major-class instances~\cite{buda2018systematic,he2009learning,japkowicz2000class,ouyang2016factors}.
\cite{AdSampling,chawla2002smote,fernandez2018smote,wang2019wgan,Chou2020Remix} proposed to synthesize more minor-class instances to enlarge the diversity of a class. More advanced methods learn to transfer statistics from the major to the minor classes~\cite{hariharan2017low,liu2019large,yin2019feature,Kim2020M2M}.

\noindent\textbf{Cost-sensitive based methods} adjust the cost of misclassifying an instance according to its true class.
One common way is to give different instance weights (\ie, the so-called re-weighting) but still apply the same instance loss function. Setting the weights by (the square roots of) the reciprocal of the number of training instances per class has been widely used~\cite{huang2016learning,huang2017discriminative,mahajan2018exploring,wang2016training,wang2017learning}. In~\cite{cui2019class}, Cui \etal proposed a principled way to set weights by computing the effective numbers of training instances for each class. Several recent methods \cite{chang2017active,jiang2018mentornet,ren2018learning,shu2019meta,wang2019dynamic,Jamal2020Rethinking,Ren2020Balanced} explored dynamic adjustments of the weights along the learning process via meta-learning or curriculum learning.

Instead of adjusting the instance weights, Khan \etal~\cite{khan2017cost} developed several instance loss functions that reflect class imbalance. Cao \etal~\cite{cao2019learning} forced minor-class instances to have large margins from the decision boundaries. Khan \etal~\cite{khan2019striking} proposed to incorporate uncertainty of instances or classes in the loss function. Ren \etal~\cite{Ren2020Balanced} and Menon \etal~\cite{menon2021longtail} modified the softmax loss function to accommodate the label distribution shift or calibrate the biased logits toward major classes.
Both Li \etal~\cite{li2020overcoming} and Tan \etal~\cite{tan2020equalization} introduced new loss functions to balance the gradients assigned to the last fully-connected layers over classes. 

Our CDT also adjusts instance loss functions, and we provide comparisons to closely related work in \autoref{ss_most_related}.

\noindent\textbf{Learning feature embeddings} with class-imbalanced data has also been studied, especially for face recognition~\cite{dong2017class,dong2018imbalanced,huang2016learning,huang2019deep,zhang2017range}. Several recent works \cite{hayat2019gaussian,wu2017deep,zhong2019unequal} combined objective functions of classifier and embedding learning to better exploit the data of minor classes. Cui \etal~\cite{cui2018large}, Zhang \etal~\cite{zhang2019study}, and Wang \etal~\cite{wang2020frustratingly} proposed two-stage training procedures to pre-train features with imbalanced data and fine-tune the classifier with balanced data. Kang \etal~\cite{kang2019decoupling} proposed to decouple a ConvNet classifier by its feature extractor and the subsequent linear classifier, and systematically studied different training strategies for each component. Interestingly, they found that re-weighting and re-sampling might hurt the feature quality. Our analysis in~\autoref{sec:analysis} provides insights to support their observations. Zhou \etal~\cite{Zhou2019BBN} introduced bilateral-branch networks to cumulatively transit between the two training stages.

\noindent\textbf{Over-fitting and under-fitting.} 
We are unaware of much recent literature that discusses if the learned classifiers under-fit minor classes, which means the classifiers also perform poorly in training~\cite{li2016solving,li2018adaptive}, or over-fit minor classes~\cite{al2016algorithms}. Cao \etal~\cite{cao2019learning} observed over-fitting when over-scaling up the influence of minor classes. We empirically show that under-fitting seems to be the main problem in traditional machine learning approaches; over-fitting seems to be prevalent in deep learning approaches trained end-to-end.

\noindent\textbf{Empirical observations.} Several works
\cite{guo2017one,kang2019decoupling,yin2019feature,Kim2020Adjusting} found that the learned linear classifiers of a ConvNet tend to have larger norms for the major classes, and proposed to force similar norms in training or calibrate the norms in testing. Wu \etal~\cite{wu2017deep} found that the feature norms of major-class and minor-class instances are different and proposed to regularize it by forcing similar norms. Our work is inspired by empirical observations as well, but from a different perspective. We find that learning with class-imbalanced data leads to feature deviation between training and test instances of the same class, which explains the problem of over-fitting to minor classes. A recent work \cite{Kim2020Adjusting} also observed the discrepancy between training and test feature distributions, from the perspective of how they affect classification boundaries. 
Our work is conducted independently from theirs\footnote{Our preliminary technical report \cite{ye2020identifying} was posted on arXiv.org in January, 2020, just a month after the arXiv version of \cite{Kim2020Adjusting}.} and is distinct by a detailed analysis of feature deviation from the perspective of over-fitting and a different way, CDT, for imbalanced deep learning.

\begin{figure*}[t]
    \begin{center}
		\begin{minipage}[h]{0.33\linewidth}
			\centering \includegraphics[width=\linewidth]{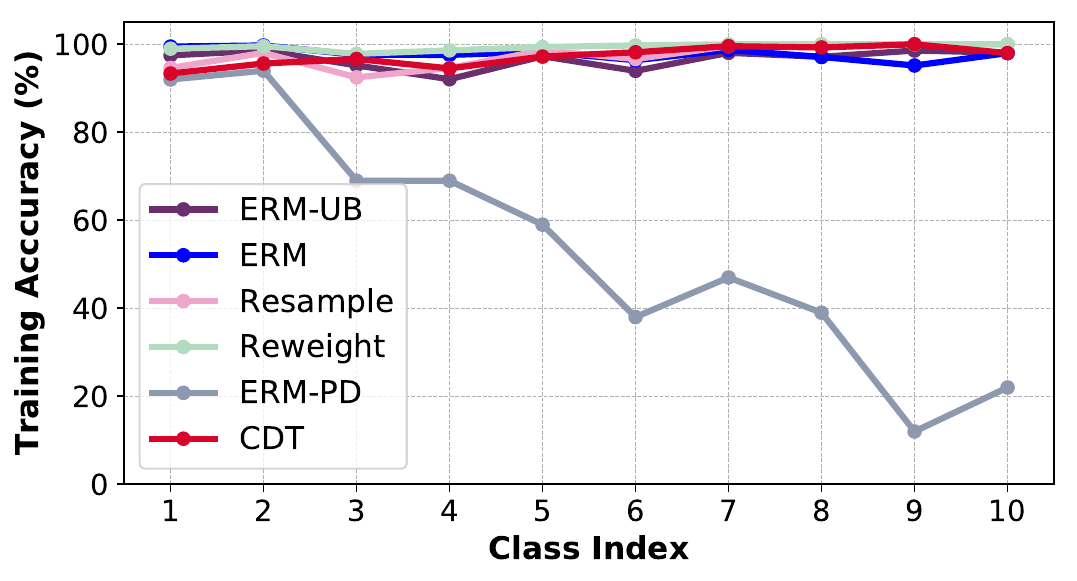}\\
			\mbox{(a) Training set accuracy}
		\end{minipage}\hfill
		\begin{minipage}[h]{0.33\linewidth}
			\centering \includegraphics[width=\linewidth]{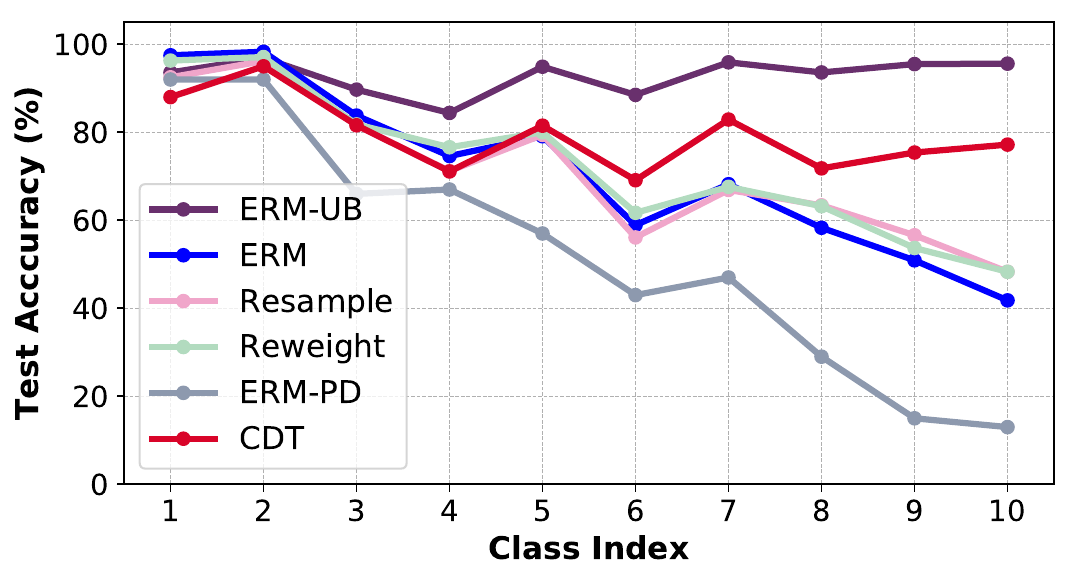}\\
					    \mbox{(b) Test set accuracy}
		\end{minipage}\hfill
		\begin{minipage}[h]{0.33\linewidth}
			\centering \includegraphics[width=\linewidth]{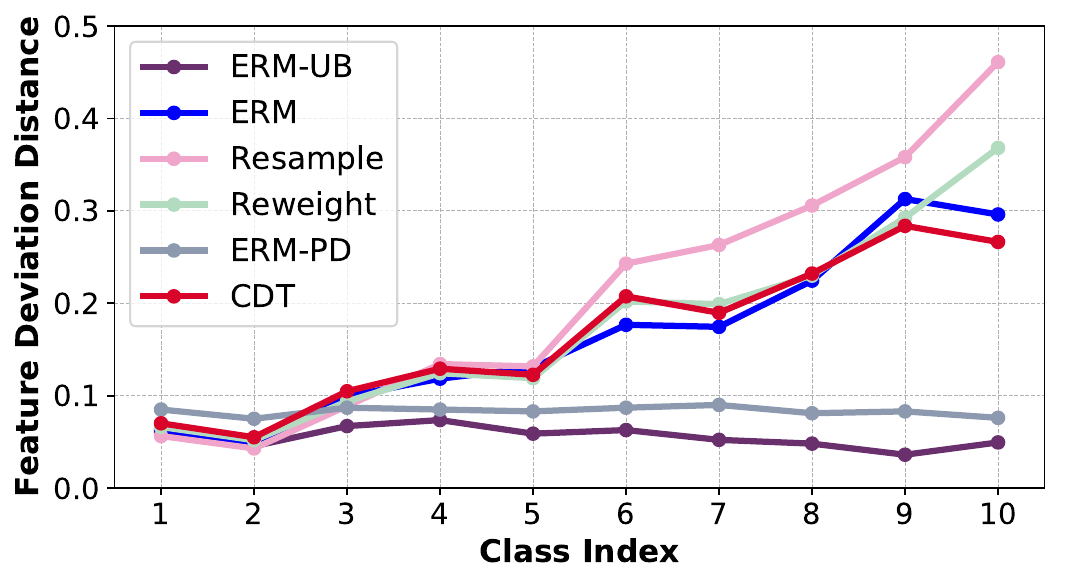}\\
					\mbox{(c) Feature deviation}
		\end{minipage}
	\end{center}
	\vskip -5pt
    \caption{\textbf{The effects of learning with class-imbalanced data.} On long-tailed CIFAR-10 \cite{cui2019class,krizhevsky2009learning} (see \autoref{fig:train_test_line_plot} (a)), we train ConvNet classifiers using ERM, re-weighting, re-sampling, and our \textbf{class-dependent temperatures (CDT)}. We also train a logistic regression using ERM and pre-defined features (denoted as ERM-PD). We further investigate the upper bound by training a ConvNet classifier using ERM on the original balanced CIFAR-10, denoted as ERM-UB. We show the (a) training set accuracy, (b) test set accuracy, and (c) feature deviation (\autoref{eq_feat_dev}) for each class. We see over-fitting to minor classes from ConvNet classifiers, but under-fitting to minor classes from logistic regression with pre-defined features. CDT can \emph{compensate for the effect of feature deviation} (but not directly reduce deviation) and thus leads to a higher test set accuracy.}
    \label{fig:2}
\end{figure*}

\noindent\textbf{Domain adaptation.} Domain adaptation~\cite{gong2012geodesic,ben2010theory,ganin2016domain,tsai2018learning,wang2020train} study the problems in which the training and test data come from different domains (\eg, synthetic vs. real images, or images captured at different weathers or geo-locations). Namely, there is an inherent domain shift even before a machine learning model is trained. In this paper, we follow the standard setup of class-imbalanced learning such that the training and test data come from the same domain (\ie, the same dataset). Thus, there is no domain shift.
This can be seen from \autoref{fig:2} (c): when data are class-balanced (ERM-UB) or features are pre-defined (ERM-PD), there is no trend of deviation. The feature deviation is caused by training ConvNets end-to-end with class-imbalanced data.


\section{The Feature Deviation Phenomenon}
\label{sec:analysis}

We present an empirical study on over-fitting to minor classes and identify the feature deviation phenomenon and its effect.

\subsection{Background and notation}
We denote a ConvNet classifier by
\begin{align}
\hat{y} = \argmax_{c\in\{1,\cdots,C\}} \vw_c^\top f_{\vtheta} (\vx), \label{eq_CNN_pred}   
\end{align}
where $\vx$ is the input, $f_{\vtheta}(\cdot)$ is the ConvNet feature extractor parameterized by $\vtheta$, and $\vw_c$ is the linear classifier of class $c$. 

Given a training set $D_\text{tr} = \{(\vx_n, y_n)\}_{n=1}^N$, in which each class $c$ has $N_c$ instances, we train the ConvNet classifier (i.e., $\vtheta$ and  $\{\vw_c\}_{c=1}^C$) by empirical risk minimization (ERM), using the cross entropy loss
\begin{align}
-\sum_n \log p(y_n|\vx_n) = -\sum_n \log\left(\frac{\exp(\vw_{y_n}^\top f_{\vtheta}(\vx_n))}{\sum_c \exp(\vw_c^\top f_{\vtheta}(\vx_n))}\right). 
\label{eq:CE}
\end{align}
We apply stochastic gradient descent (SGD) with uniformly sampled instances from $D_\text{tr}$.

\subsection{Learning with class-imbalanced data}
We follow~\cite{cui2019class} to construct a long-tailed class-imbalanced $D_\text{tr}$ from CIFAR-10~\cite{krizhevsky2009learning}: the (most) major class contains 5,000 training instances while the (most) minor class contains 50 training instances. Without loss of generality, let us re-index classes so that the training instances decrease from class $c=1$ to $c=10$. See~\autoref{fig:train_test_line_plot} (a) for an illustration of the data size. We learn a ConvNet classifier using the standard ResNet-32 architecture and learning rates~\cite{he2016deep}, following~\cite{cao2019learning,cui2019class}. See \autoref{sec:setup} for more details on the setup.

\autoref{fig:2} (a-b, \textbf{ERM} curves) shows the per-class training and test accuracy. The training accuracy is high ($\sim 100\%$)~\cite{zhang2016understanding} for all classes. The test accuracy, however, drops drastically for the minor classes. The learned classifier over-fits minor class data. (For comparison, we train a classifier using the entire class-balanced CIFAR-10. See the \textbf{ERM-UB} curves.)

\subsection{Analysis on the effect of feature deviation}
\label{ssec:analysis}
We find that the learned linear classifiers have larger norms (i.e., $\|\vw_c\|_2$) for the major classes than the minor classes, which confirms the observations in~\cite{guo2017one,kang2019decoupling,khan2019striking,yin2019feature,Kim2020Adjusting}. The ConvNet classifier is thus biased to classify instances into major classes. \emph{This observation alone, however, cannot explain over-fitting. Specifically,
with the biased classifier norms toward major classes, why can a ConvNet classifier still achieve $\sim 100\%$ accuracy for minor-class training data, but not for test data?}

\noindent\textbf{Feature deviation.} We hypothesize that there are differences in the features (i.e., $f_{\vtheta}(\vx)$) between the training and test instances, especially for minor classes.
To verify this, we compute the feature means of training and test data for each class $c$, and then calculate their Euclidean distance $dis(c)$
\begin{align}
    & dis(c) = \nonumber \\ & \frac{1}{R}\sum_{r=1}^R \|\text{mean}(S_K(\{f_{\vtheta}(\vx_{\text{train}}^{(c)})\}))-\text{mean}(\{f_{\vtheta}(\vx_{\text{test}}^{(c)})\})\|_2. \label{eq_feat_dev}
\end{align}
Here, $\{f_{\vtheta}(\vx_{\text{train}}^{(c)})\}$ and $\{f_{\vtheta}(\vx_{\text{test}}^{(c)})\}$ denote the sets of feature vectors extracted from the training and test data of class $c$. $S_K(\cdot)$ is an operation that sub-samples $K$ instances uniformly, for $R$ rounds\footnote{We normalize a feature vector to unit $\ell_2$ norm, following~\cite{wang2019simpleshot}. We perform sub-sampling to ensure a fair comparison among classes --- even for the same class, using fewer data to compute the mean will suffer a higher variance. By using the same number of data to compute the means, the resulting $dis(c)$ could better reflect feature deviation for each class $c$. We set $K$ to be  the minor-class size and set $R=1,000$.}.
\autoref{fig:2} (c) shows the results: $dis(c)$ goes large as the number of training instances decreases. In other words, the features $f_{\vtheta}(\vx)$ of the training and test instances are deviated (e.g., translated) for minor classes. See~\autoref{fig1:feature_d} for the t-SNE~\cite{maaten2009visualizing} visualization of the features.

\begin{figure*}[t]
    \centering
    \begin{tabular}{@{\;}c@{\;}c@{\;}}
    	\begin{minipage}[h]{0.48\textwidth}
    		\centering
    	    \includegraphics[width=\textwidth]{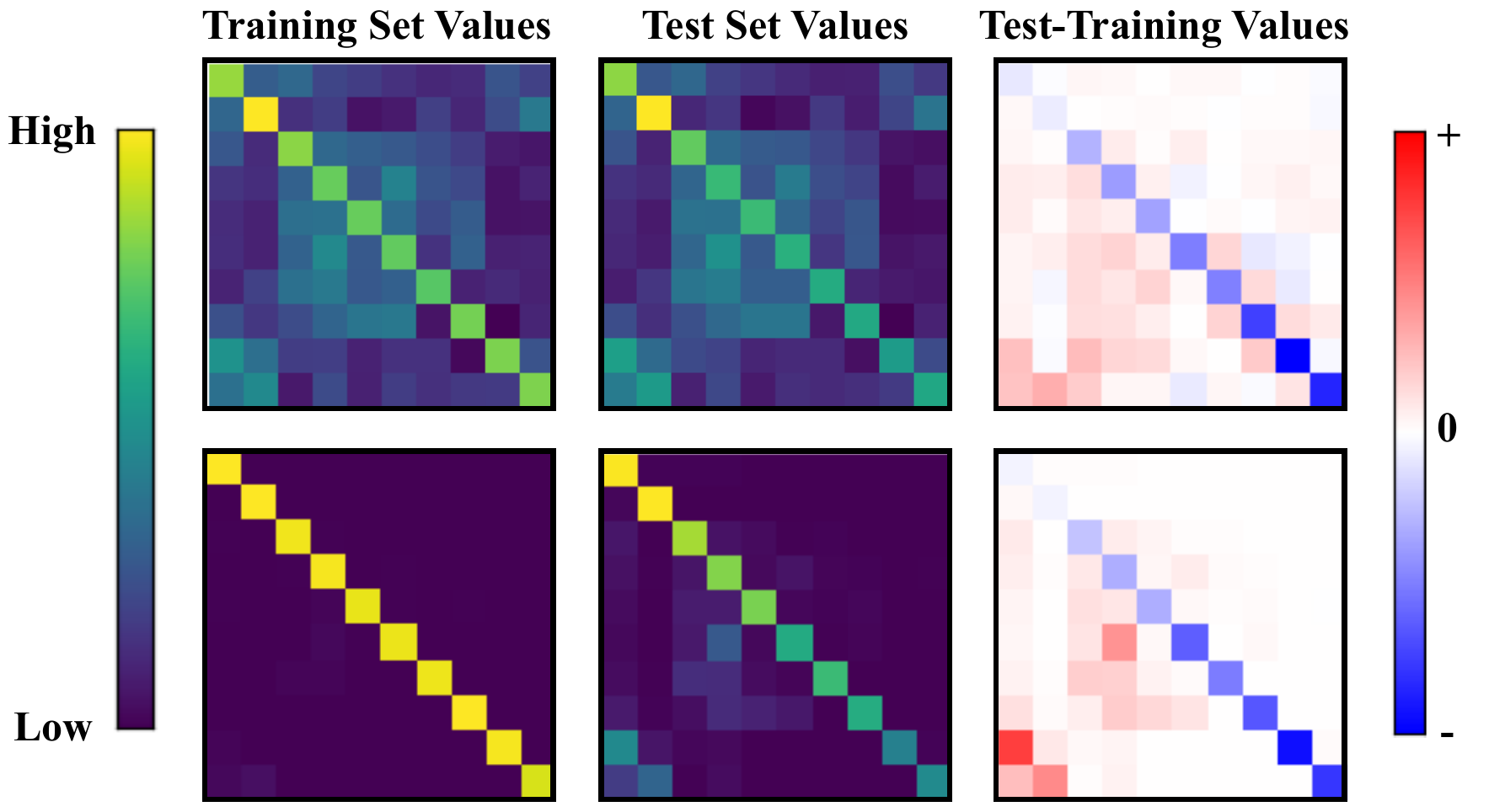}
    		\\\mbox{(a) ERM}
    	\end{minipage} & 
    	\begin{minipage}[h]{0.48\textwidth}
    		\centering	\includegraphics[width=\textwidth]{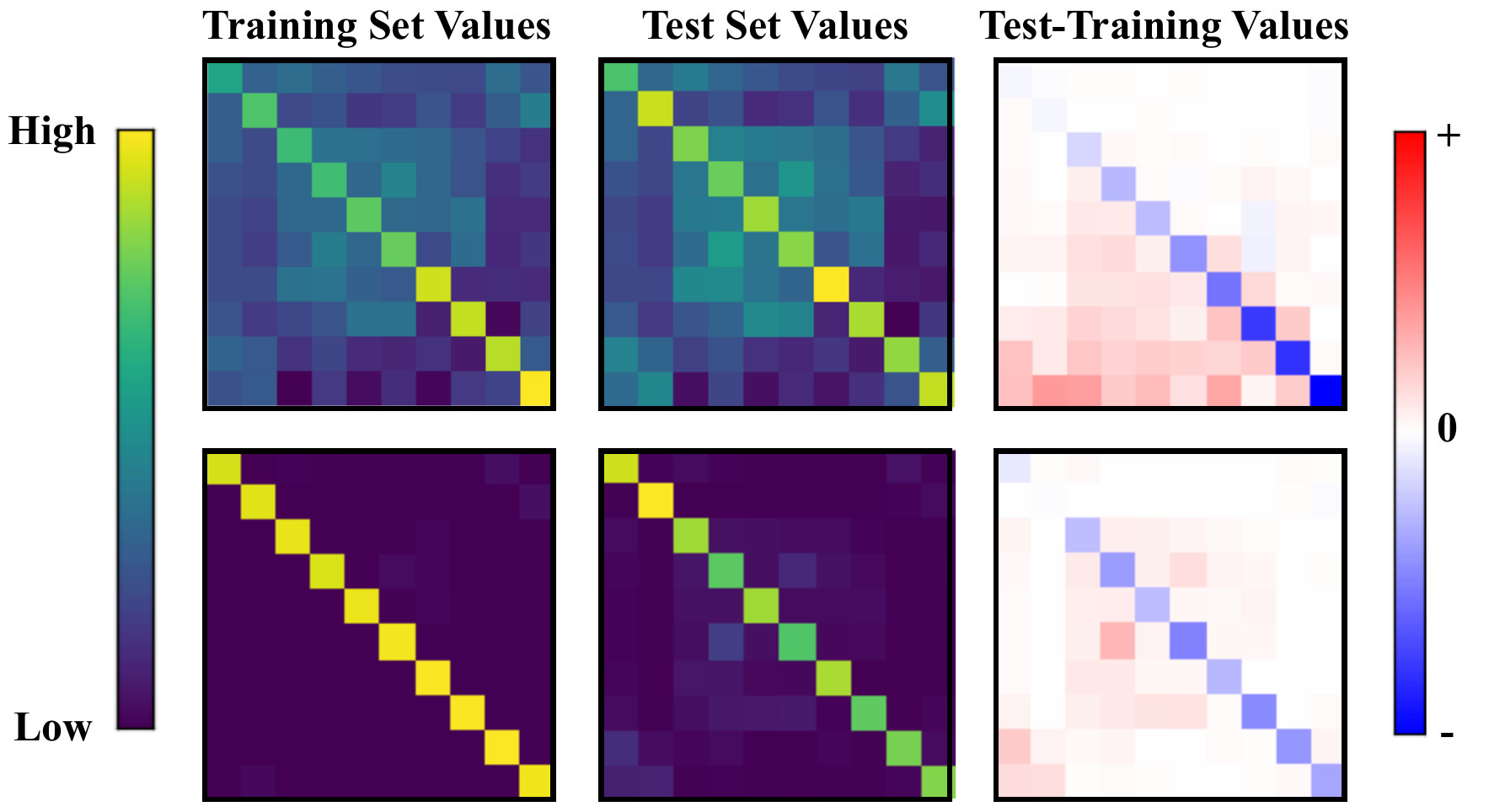}\\\mbox{(b) CDT (ours)}
    	\end{minipage}
    \end{tabular}
    \caption{Confusion matrices of \textbf{decision values (top)} and \textbf{predicted labels (bottom)} in the training (left) and test (middle) sets, for a ConvNet trained with ERM \textbf{(a)} and CDT \textbf{(b)}. The experimental setup follows \autoref{fig:train_test_line_plot}. Each row/column in a matrix corresponds to the true/predicted label.
    The numbers of training data decrease from $c=1$ (\ie, row 1) to $c=10$.}
    \label{fig:3}
\end{figure*}

The \emph{feature deviation} phenomenon offers an explanation to over-fitting to minor classes. Concretely, the linear classifiers $\{\vw_c\}_{c=1}^C$ learned to separate the training instances among classes (almost perfectly) have little to do with separating the test instances for minor classes, since the training and test features of the same minor class are not occupying the same region of the feature space. 

To further understand how feature deviation correlates with over-fitting, we plot the confusion matrices in \autoref{fig:3} (a).
For data of each class $c'$ (row-wise), we show the decision values $\vw_{c}^\top f_{\vtheta}(\vx)$ and the predictions\footnote{The prediction is the $\argmin$ of decision values (\ie, logits) over all possible labels (see \autoref{eq_CNN_pred}).} to all the possible labels $c$ (column-wise), averaged over the training or test instances. Comparing the matrix in testing (middle) to that in training (left)\footnote{The right sub-figure in ~\autoref{fig:3} (a) or (b) shows the difference.}, we have the following findings.
\begin{itemize}
    \item The diagonal decision values drop (get darker) in testing, especially for minor classes, suggesting that \emph{the deviated features of test instances tend to move to regions of lower decision values.} This can be seen in the right sub-figure, which shows the difference between testing and training. We note that, the diagonal decision values are similar across classes in training, even though the classifier norms are biased to major classes.
    \item The off-diagonal decision values do not change much from training to testing, compared to the diagonal values. Some off-diagonal values even increase, especially those at the lower triangle, meaning that \emph{the deviated features of minor classes tend to move toward major classes.}
\end{itemize}

These effects of feature deviation explain why a ConvNet classifier misclassifies many of the minor-class test instances into major classes, as shown in the confusion matrix of predicted labels in~\autoref{fig:3} (a), the bottom middle sub-figure.

\subsection{Scaling up minor classes increases deviation}
\label{ssec:scaling_up}
We investigate how popular treatments such as re-weighting and re-sampling affect ConvNet learning. Let $N_c$ be the number of training instances of class $c$, we re-weight an instance of class $c$ by $N_c^{-0.5}$, as suggested in~\cite{maaten2009visualizing}. We consider re-sampling so that every mini-batch has the same number of instances per class.
\autoref{fig:2} shows the resulting training and test set accuracy and the distance between training and test feature means per class.
Both methods do not reduce the feature mean distance (\ie, feature deviation) or even aggravate it. The test accuracy thus remains poor.

To better understand feature deviation, we follow~\cite{kang2019decoupling} to apply a nearest center classifier using the training feature means as centers.
\autoref{tab:NCM} shows the results. We see a similar trend as in~\cite{kang2019decoupling}: ERM outperforms re-sampling and is on a par with re-weighting. {The average test set accuracy is even higher than applying the learned linear classifiers.} We attribute the relatively poor performance of re-sampling to the feature mean deviation, since we use the training means as centers. We further investigate the upper bound by using the test feature means as centers (thus removing the deviation of means). 
Re-sampling still falls behind others, suggesting that it also degrades the quality (e.g., discriminative ability) of features.
In short, our observations suggest that naively scaling up the influence of minor classes in training might aggravate feature deviation (or reduce feature quality) and can not resolve problems in class-imbalanced learning.

\begin{table}[t]
	\centering
	\small
	\caption{Average test set accuracy using ConvNet features with the learned linear classifiers or nearest center classifiers on imbalanced CIFAR-10 (see \autoref{fig:train_test_line_plot} (a)). We compare using training and test feature means (the upper bound) as class centers. RW: re-weighting; RS: re-sampling; CDT: our approach in~\autoref{sec:approach}.
	}
	\label{tab:NCM}
	\begin{tabular}{l|c|c|c|c}
		\toprule
		Method & ERM & RS & RW & \textbf{CDT} \\ \hline
		Linear classifiers & 71.1 & 71.2 & 72.6 & 79.4 \\
		Training means & 77.0 & 72.3 & 77.0 & 78.8	\\ 
		Test means (upper bound) & 79.5 & 75.8 & 79.3 & 79.4\\
		\bottomrule
	\end{tabular}
\end{table}

\subsection{Traditional vs. end-to-end training}
\label{ssec:traditional_deep}
We have shown that ConvNet classifiers trained end-to-end with class-imbalanced data suffers feature deviation and over-fitting to minor classes.
In traditional machine learning, features are usually pre-defined and are not learned from the training data at hand. 
\emph{In this scenario, will feature deviation and over-fitting still occur?}
To answer this, we investigate training a ConvNet classifier using another dataset (we use Tiny-ImageNet~\cite{le2015tiny}). We apply the learned feature extractor $f_{\vtheta}(\cdot)$ to the imbalanced CIFAR-10, for both training and test instances. We then train a logistic regression classifier using features of the training instances and apply it to the test instances. \autoref{fig:2} (\textbf{ERM-PD}) shows the results. We see \emph{no trend of feature deviation}. The resulting classifier performs poorly on the minor classes, not only in testing but also in training, essentially suffering \emph{under-fitting to minor classes.}

We thus argue that, while learning with class-imbalanced data hurts the accuracy on minor classes for both traditional machine learning and recent deep learning, the underlying reasons are drastically different. Indeed, we show that re-weighting is effective to alleviate under-fitting for logistic regression (in~\autoref{ssec:linear}) but not for ConvNet which suffers over-fitting.
It is therefore desirable to develop solutions for a ConvNet classifier (or other deep classifiers) by incorporating the insights from feature deviation.

\section{Class-Dependent Temperatures (CDT)}
\label{sec:approach}

We present a novel approach to learn a ConvNet classifier with class-imbalanced data, inspired by our observations of feature deviation. Specifically, according to \autoref{ssec:analysis}, feature deviation makes the diagonal decision values of minor classes (see \autoref{fig:3} (a)) drop significantly from the training data to the test data. For test instances of some minor classes (e.g., class $c=9$ and $c=10$), the diagonal values are close to or even lower than some of the off-diagonal values, making it hard to classify them correctly.

One way to remedy this issue is to prevent feature deviation, which, however, appears difficult based on our earlier studies. We therefore take an alternative way, seeking to design a training objective that can \emph{compensate for the effect of feature deviation} --- \ie, compensate for the reduced diagonal decision values of minor classes in testing --- so that the learned classifier can classify minor class data correctly.

To this end, we propose to \emph{make the training process aware of feature deviation, by simulating the effect of feature deviation directly on the training instances.} Observing that feature deviation decreases decision values, we \emph{artificially} reduce the decision values $\vw_c^\top f_{\vtheta}(\vx)$ of a training instance by a factor (temperature) $a_c$, resulting in a new objective
\begin{align}
-\sum_n \log\left(\frac{\exp\left(\cfrac{\vw_{y_n}^\top f_{\vtheta}(\vx_n)}{a_{y_n}}\right)}{\sum_c \exp\left(\cfrac{\vw_c^\top f_{\vtheta}(\vx_n)}{a_c}\right)}\right). 
\label{eq:margin}
\end{align}
We set $a_c$ inversely proportionally to the number of training instances $N_c$ --- i.e., minor classes have larger $a_c$ --- reflecting the fact that minor classes suffer larger feature deviation. Training a ConvNet classifier to minimize \autoref{eq:margin} therefore forces $\vw_c^\top f_{\vtheta} (\vx)$ to be enlarged by a factor of $a_c$ for $\vx$ belonging to class $c$, so as to compensate for the effect of feature deviation that will occur on the test data.

Without loss of generality, we set $a_c = \left(\frac{N_{\max}}{N_c}\right)^\gamma$, in which $\gamma\geq 0$ is a hyperparameter and $N_{\max}$ is the largest number of training instances of a class in $D_\text{tr}$. That is, $a_c=1$ for the (most) major class and $a_c>1$ for the other classes if $\gamma>0$. When $\gamma=0$, we recover the conventional ERM objective. We note that, it is quite common to set a class-dependent coefficient proportionally or inversely proportionally to $N_c^{\gamma}$ \cite{buda2018systematic,byrd2019what,cao2019learning,huang2016learning,Kim2020Adjusting,mahajan2018exploring} (with a different $\gamma$).

\noindent\textbf{Testing.} CDT \emph{simulates} the effect of feature deviation on the training data, which encourages the ConvNet classifier to learn to deal with it by generating larger decision values for minor class data. In the test phase, we apply the same classification rule as in \autoref{eq_CNN_pred},
\begin{align}
\hat{y} = \argmax_{c\in\{1,\cdots,C\}} \vw_c^\top f_{\vtheta} (\vx). \label{eq_CNN_pred_CDT}
\end{align}
Namely, we remove the factor $a_c$. This is because the test data themselves suffer \emph{real} feature deviation, and hence there is no need to further \emph{simulate} feature deviation on them.

\subsection{Analysis on CDT}
\autoref{fig:3} (b) shows the confusion matrices of applying our learned ConvNet classifier with CDT ($\gamma = 0.3$) to the training and test instances, using~\autoref{eq_CNN_pred_CDT}.  The experimental setup follows \autoref{fig:train_test_line_plot}. On the training data, minor classes now have larger diagonal decision values than major classes. 
On the test data, due to the unavoidable feature deviation, the diagonal decision values drop for minor classes in comparison to those on the training data. However, the diagonal values of all classes are much balanced than in~\autoref{fig:3} (a). As a result, the test accuracy increases for not only minor classes but also all classes on average (see \autoref{tab:NCM} and \autoref{fig:2}).

\subsection{Discussion and comparison to related work}
\label{ss_most_related}
Our approach shares similar influences to the classifier's decision values with~\cite{cao2019learning,Ren2020Balanced,menon2021longtail}. However, our motivations are quite different --- \emph{we conduct analysis to identify feature deviation and its effect and propose CDT to compensate for it.} In contrast, \cite{cao2019learning} is proposed according to a margin-based generalization bound, \cite{Ren2020Balanced} is proposed to accommodate the label distribution shift between training and testing, and \cite{menon2021longtail} is proposed to overcome the undesirable bias towards dominant labels. Our ways of adjusting the decision values are also different from \cite{cao2019learning,Ren2020Balanced,menon2021longtail}, as summarized in a recent comprehensive study~\cite{wu2021adversarial}.
We therefore believe that our insights and approach are complementary to them and are valuable add-on to the community\footnote{Our preliminary technical report \cite{ye2020identifying} has been cited in \cite{Ren2020Balanced,menon2021longtail}.}.

\section{Experimental Setups}
\label{sec:setup}

\subsection{Datasets} We experiment on four datasets, namely CIFAR-10~\cite{krizhevsky2009learning}, CIFAR-100~\cite{krizhevsky2009learning}, Tiny-ImageNet~\cite{le2015tiny}, and iNaturalist (2018 version)~\cite{van2018inaturalist}. \textbf{CIFAR-10} and \textbf{CIFAR-100}~\cite{krizhevsky2009learning} are balanced image classification datasets composed of $50,000$ training and $10,000$ test images of 32 $\times$ 32 pixels from 10 and 100 classes, respectively. \textbf{Tiny-ImageNet}~\cite{le2015tiny} has 200 classes; each class has 500 training and 50 validation images of $64\times 64$ pixels. \textbf{iNaturalist}~\cite{van2018inaturalist} (2018 version) is an imbalanced long-tailed dataset which contains 437,513 training images from 8,142 classes and 3 validation images per class.  

\begin{figure*}[t]
	\minipage{0.24\textwidth}
	\includegraphics[width=1.\linewidth]{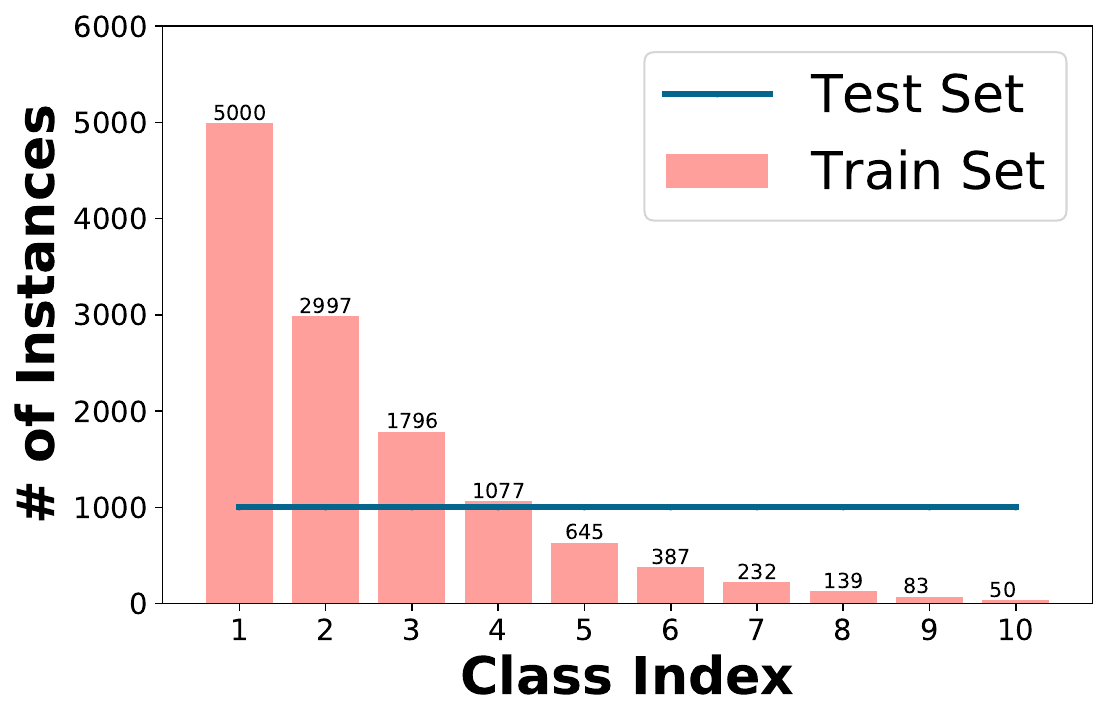}
	\centering
	\mbox{\footnotesize CIFAR-10, LT ($\rho$=100)}
	\endminipage\hspace{0.05cm}
	\minipage{0.24\textwidth}
	\includegraphics[width=1.\linewidth]{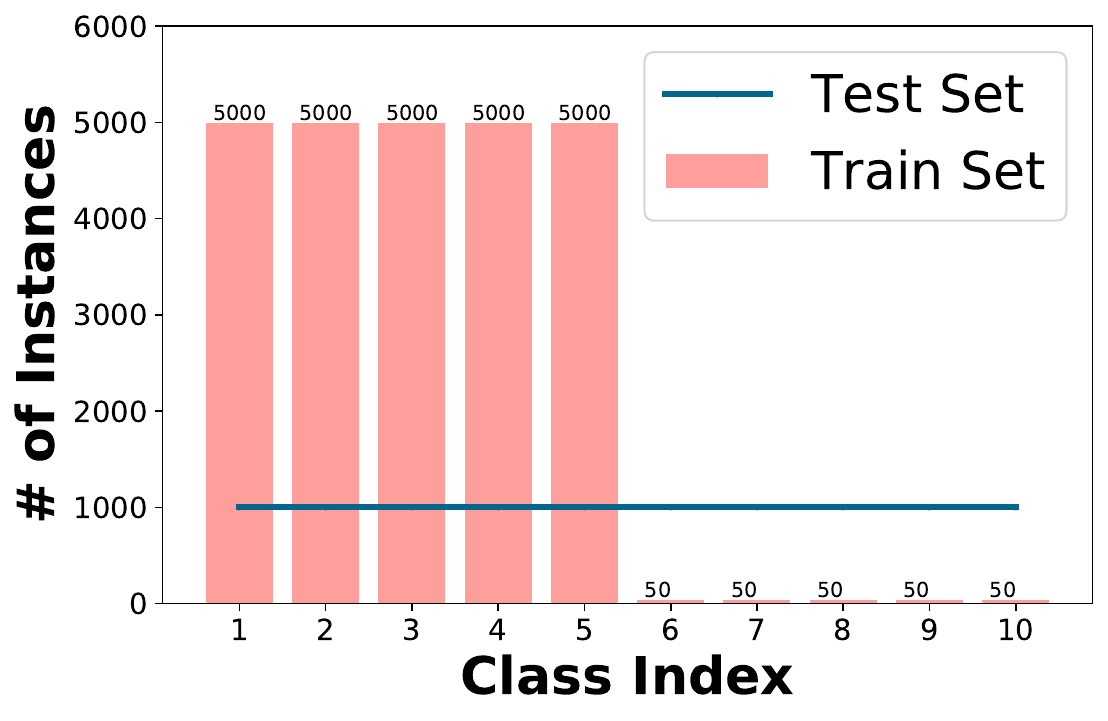}
	\centering
	\mbox{\footnotesize CIFAR-10, step ($\rho$=100)}
	\endminipage\hspace{0.05cm}
	\minipage{0.24\textwidth}
	\includegraphics[width=1.\linewidth]{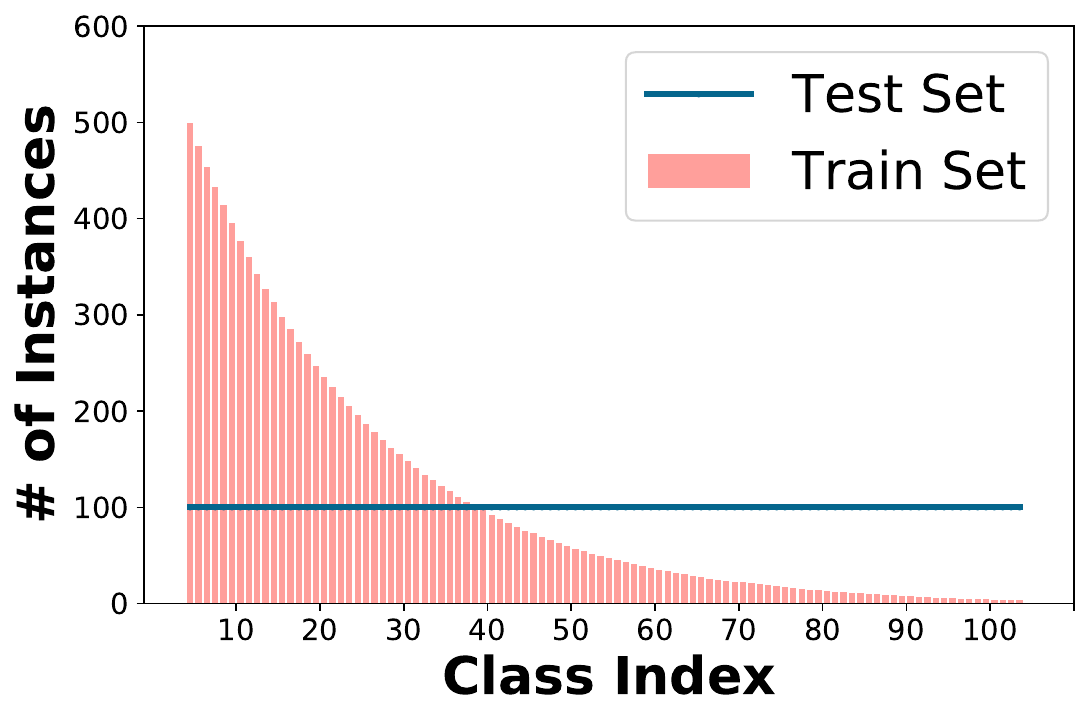}
	\centering
	\mbox{\footnotesize CIFAR-100, LT ($\rho$=100)}
	\endminipage\hspace{0.05cm}
	\minipage{0.24\textwidth}
	\includegraphics[width=1.\linewidth]{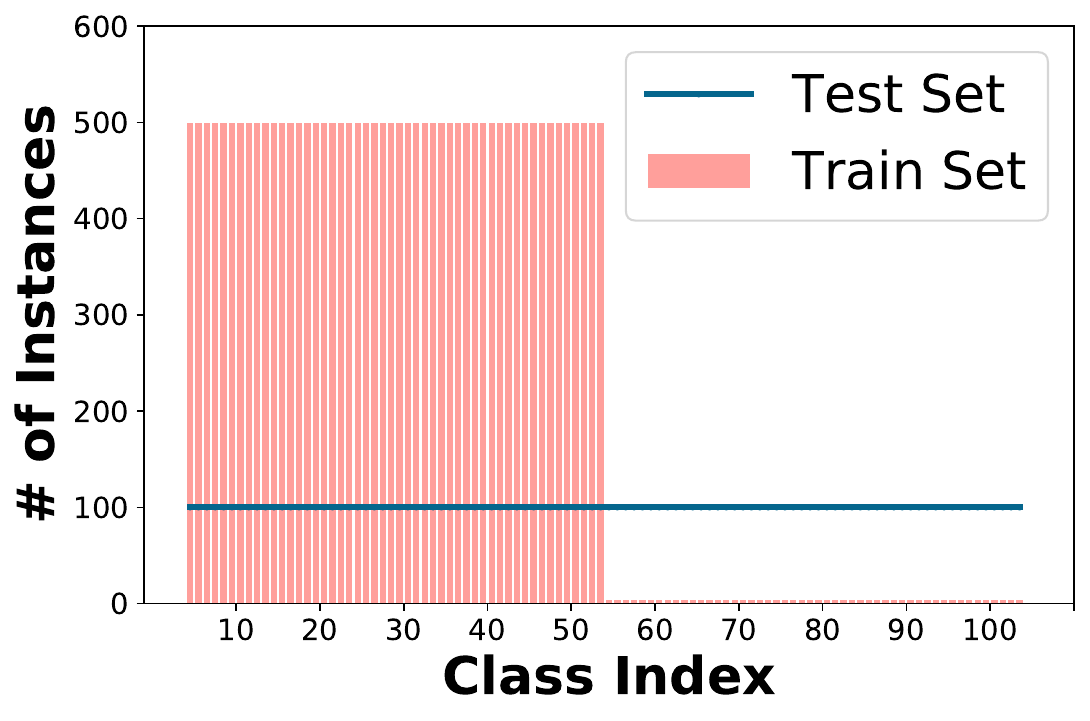}
	\centering
	\mbox{\footnotesize CIFAR-100, step ($\rho$=100)}
	\endminipage\hspace{0.05cm}
	\minipage{0.24\textwidth}
	\includegraphics[width=1.\linewidth]{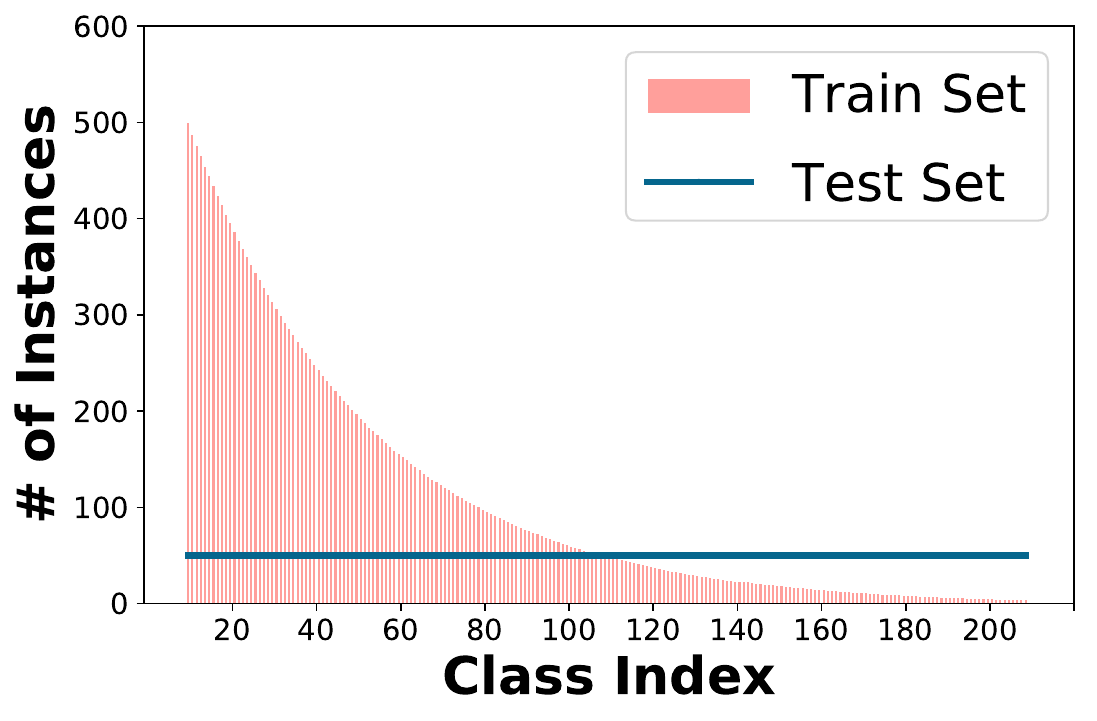}
	\centering
	\mbox{\footnotesize T-ImageNet, LT ($\rho$=100)}
	\endminipage\hspace{0.05cm}
	\minipage{0.24\textwidth}
	\includegraphics[width=1.\linewidth]{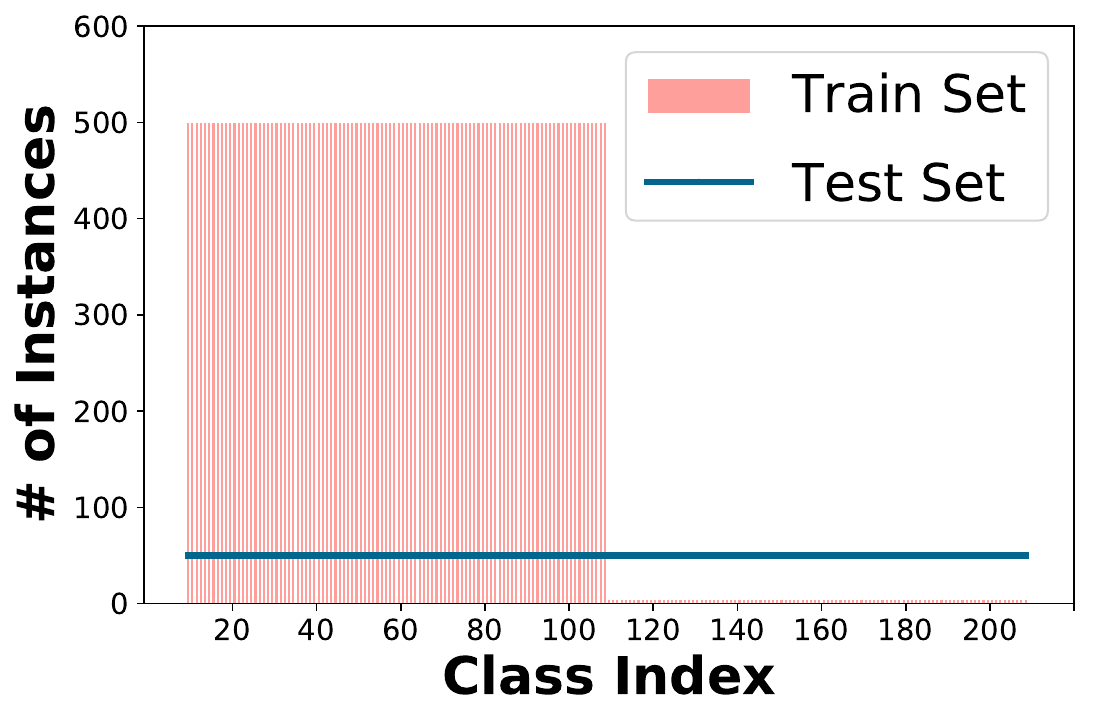}
	\centering
	\mbox{\footnotesize T-ImageNet, step ($\rho$=100)}
	\endminipage\hspace{0.05cm}
	\minipage{0.24\textwidth}
	\includegraphics[width=1.\linewidth]{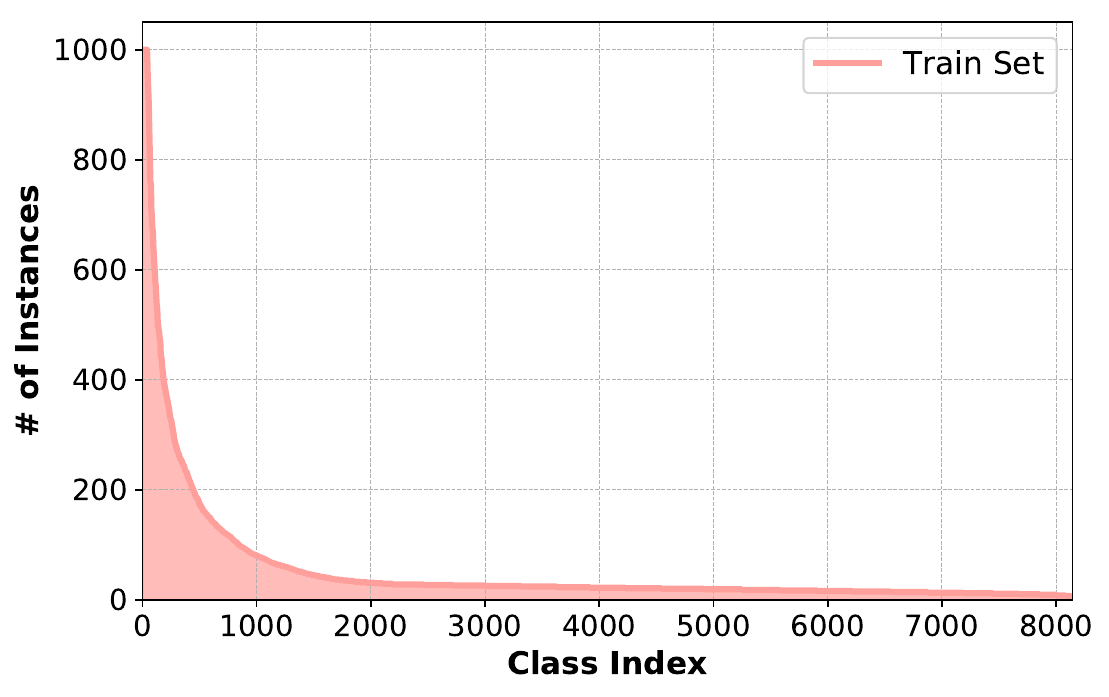}
	\centering
	\mbox{\footnotesize iNaturalist}
	\endminipage
	\caption{\textbf{Dataset Illustrations.} The numbers of training and test (or validation) instances per class are indicated by red and blue color. We do not show the validation sets of iNaturalist since each class only has 3 instances. LT: long-tailed; T-ImageNet: Tiny-ImageNet.}
	\label{fig:dataset}
\end{figure*}

The original CIFAR-10, CIFAR-100, and Tiny-ImageNet are balanced, and we follow the strategy in \cite{cao2019learning,cui2019class} to create class-imbalanced training set with an imbalance ratio, such that the numbers of instances per class follow a certain distribution. 
In our experiments we consider two distributions: \emph{long-tailed (LT)} imbalance follows an exponential distribution; \emph{step} imbalance reduces training instances of half of the classes to a fixed size. We control the degree of dataset imbalance by the imbalance ratio $\rho=\frac{N_\text{max}}{N_\text{min}}\in\{10, 100, 200\}$, where $N_\text{max}$ ($N_\text{min}$) is the number of training instances of the largest major (smallest minor) class.
The test and validation sets remain unchanged and balanced. We re-index classes so that smaller indices have more training instances.

Following existing works, we report the accuracy of CIFAR-10 and CIFAR-100 on the test set, and report the accuracy of Tiny-ImageNet and iNaturalist on the validation set. See \autoref{fig:dataset} for illustrations of some datasets.


\subsection{Implementation Details} 
\noindent\textbf{ERM and class-dependent temperatures (CDT).} For all the experiments, we use mini-batch stochastic gradient descent (SGD) with momentum = 0.9 as the optimization solver. The softmax cross-entropy loss (\autoref{eq:CE}) is used for ERM and our CDT. The weight decay is $2 \times 10^{-4}$. 

We also incorporate the Deferred Re-Weighting (DRW) \cite{cao2019learning} in the training phase, which is orthogonal to CDT and usually introduces further improvements. DRW turns to minimizes a re-weighted objective after 60\% training epochs, in which the minor-class instances have larger weights. We follow~\cite{cao2019learning} to set the weights according to~\cite{cui2019class}. The weight for instances of class $c$ is $\frac{1-\beta}{1-\beta^{N_c}}$ with $\beta=0.9999$. 
	
\noindent\textbf{Training details for imbalanced CIFAR.} We use ResNet-32~\cite{he2016deep} for all the CIFAR experiments, following~\cite{cao2019learning, cui2019class}.
The batch size is 128. The initial learning rate is linearly warmed up to $0.1$ in the first 5 epochs, and decays at the 160th and the 180th epochs by $0.01$, respectively. The model is trained for 200 epochs by default. For some experiments with the step imbalance, we found that 200 epochs are not enough for the model to converge and we train for 300 epochs in total and adjust the learning rate schedule accordingly. We follow \cite{cao2019learning} to perform data augmentation. The $32 \times 32$ CIFAR images are padded to $40 \times 40$ and randomly flipped horizontally, and then are randomly cropped back to $32 \times 32$. 

We also additionally consider the training protocol proposed in \cite{Ren2020Balanced}, where strong augmentations like RandAugment~\cite{Cubuk2020Rand} are used in the training phase. In this case, larger batch size (512) and longer training epochs (500) are used. We also decrease the learning rate from 0.2 with a cosine schedule. The improved training strategy leads to higher performance for both ERM and CDT.

\noindent\textbf{Training details for imbalanced Tiny-ImageNet.} We use ResNet-18~\cite{he2016deep} for Tiny-ImageNet, following~\cite{cao2019learning}. It is trained for 400 epochs with a batch size of 128. The initial learning rate is 0.1 and decays by 0.1 at the 250th, 320th, and 380th epochs. Images are padded 8 pixels on each side and randomly flipped horizontally, and are randomly cropped back to $64 \times 64$.

\noindent\textbf{Training details for iNaturalist.} We follow~\cite{cui2019class,Zhou2019BBN} to use ResNet-50~\cite{he2016deep} for iNaturalist. We train the model for 180 epochs with a batch size of 128. 
The learning rate is 0.05 at first. It decays at the 120th and the 160th by 0.1.
For pre-processing and data augmentation, we follow \cite{he2016deep,Zhou2019BBN}. We normalize the images by subtracting the RGB means computed on the training set. In training, the images are resized to $256 \times 256$ and flipped horizontally, and are randomly cropped back to $224 \times 224$.

\subsection{Hyperparameter $\gamma$}
The only hyperparameter that we need to specifically tune for CDT is $\gamma$, which defines the temperature $a_c = \left(\frac{N_{\max}}{N_c}\right)^\gamma$ in \autoref{eq:margin}.
As will be shown in \autoref{fig:gamma}, increasing $\gamma$ eventually reduces the \emph{training} accuracy of major classes. We use this property to set a range of $\gamma$ for different experiments. For CIFAR-10, $\gamma\in [0.0, 0.4]$. For CIFAR-100 and Tiny-ImageNet, $\gamma\in [0.0, 0.2]$. For iNaturalist, $\gamma\in [0.0, 0.1]$. We then select $\gamma$ using a held-out validation set from the training data (we hold out $K=3$ instances per class). We remove classes with fewer than $K$ training instances in this process.

\subsection{Baselines}
We compare to learning a ConvNet classifier using vanilla \textbf{ERM} with the cross-entropy loss, \textbf{re-sampling} in which every mini-batch has the same number of instances per class, and \textbf{re-weighting} in which every instance has a weight $N_c^{-\gamma}$ ($\gamma$ is a hyperparameter). We also compare to re-weighting with the \textbf{focal loss}~\cite{lin2017focal}, re-weighting with the \textbf{CB loss}~\cite{cui2019class},
\textbf{LDAM}~\cite{cao2019learning} with the DRW scheduling, bilateral-branch networks (\textbf{BBN})~\cite{Zhou2019BBN}, \textbf{$\tau$-norm} and learnable weight scaling (\textbf{LWS})~\cite{kang2019decoupling}, \textbf{ReMix}~\cite{Chou2020Remix}, \textbf{BALMS}~\cite{Ren2020Balanced}, logit adjustment (\textbf{LogitAdjust})~\cite{menon2021longtail}, and rethinking from the domain adaptation perspective (abbreviated as \textbf{Rethinking})~\cite{Jamal2020Rethinking}.

\section{Experimental Results}\label{sec:exp}
\subsection{Results on CIFAR-10 and CIFAR-100}
\label{sec:cifar100}

We extensively experiment with CIFAR-10 and CIFAR-100 with imbalance ratios $\rho\in\{10, 100, 200\}$. The results on long-tailed imbalance and step imbalance as shown in \autoref{tab:cifar-table} and \autoref{tab:cifar-table-step}, respectively. (We note that, some baselines do not report results on step imbalance.) Using the same training strategy as in~\cite{cao2019learning,cui2019class}, our CDT outperforms baselines in many of the settings. 
We also investigate the training strategy in~\cite{Ren2020Balanced}, in which all methods including the ERM get higher accuracy. With this strategy, CDT still get the best or competitive results. 
When combined with DRW, the accuracy of CDT can be further improved in some cases. Importantly, compared to \cite{cao2019learning,menon2021longtail,Ren2020Balanced} which share similar algorithm designs with ours (see \autoref{ss_most_related}), we attain higher accuracy in most of the settings.

\begin{table}[t]
		\small
		\tabcolsep 4pt
	\caption{\textbf{Test accuracy (\%) on \emph{long-tailed} CIFAR-10/-100}. We show reported results, unless stated otherwise. 
	$\dagger$: our reproduced results. $^\sharp$: best reported results taken from~\cite{cao2019learning,cui2019class}. Two training strategies for CIFAR are explored in the upper~\cite{cao2019learning,cui2019class} and bottom~\cite{Ren2020Balanced} blocks, and we denote the best performance using each strategy in bold.
	}
	\centering
	\begin{tabular}{l@|ccc|ccc}
		\toprule
		Dataset           & \multicolumn{3}{c|}{ CIFAR-10}                               & \multicolumn{3}{c}{ CIFAR-100}                              \\ \midrule
		Imbalance Ratio        & \multicolumn{1}{c}{200} & 100  & 10 & \multicolumn{1}{c}{200} & 100 & 10    \\ 
		\midrule
	ERM~$\dagger$  & 65.6 & 71.1 & 87.2 & 35.9 & 40.1 & 56.9 \\
    Re-sampling~$\dagger$  & 64.4  & 71.2  & 86.5  & 30.6  & 34.7  & 54.2 \\
    Re-weighting~$\dagger$  & 68.6  & 72.6  & 87.1  & 35.0  & 40.5  & 57.3   \\
    Focal~\cite{lin2017focal}~$^\sharp$ & 65.3  & 70.4  & 86.8 &  35.6  & 38.7  & 55.8   \\
    CB~\cite{cui2019class}~$^\sharp$ & 68.9  & 74.6  & 87.5  & 36.2  & 39.6  & 58.0 
    \\
    LDAM~\cite{cao2019learning}  & 69.4$\dagger$  & 73.4  & 87.0  & 36.7$\dagger$  & 39.6  & 56.9   \\
    LDAM + DRW~\cite{cao2019learning} & 74.6$\dagger$  & 77.0  & 88.2   & 39.5$\dagger$ & 42.0  & 58.7 \\
    $\tau$-norm~\cite{kang2019decoupling}~$\dagger$ & 70.3 & 75.1 & 87.8 & 39.3 & 43.6 & 57.4  \\
    BBN~\cite{Zhou2019BBN} & -  & 79.8  & 88.3   & - & 42.6  & 59.1 \\
    ReMix~\cite{Chou2020Remix} & -  & 75.4  & 88.2   & - & 41.9  & 59.4 \\
    ReMix + DRW ~\cite{Chou2020Remix} & -  & 79.8  & 89.0   & - & \bf 46.7  & \bf 61.2 \\
    Rethinking~\cite{Jamal2020Rethinking} & 77.2  & 80.0  & 87.4   & 39.5 & 44.1  & 58.0 \\
    LogitAdjust~\cite{menon2021longtail} & -  & 77.6  & -   & - & 43.9  & - \\
    \midrule 
    \textbf{CDT} \textbf{(Ours)}  & {74.7} & 79.4 & {\bf 89.4} & 40.5 & {44.3} & 58.9 \\
   	\textbf{CDT} \textbf{(Ours) + DRW}  & {\bf 77.5} & \bf 80.4 & {\bf 89.4} & \bf 41.6 & {\bf 46.7} & 60.0 \\
   	\midrule
   	\midrule
   	ERM~\cite{Ren2020Balanced}  & 70.9 & 79.1 & 91.1 & 41.5 & 45.9 & 63.6 \\
   	B-Softmax~\cite{Ren2020Balanced}  & 79.0 & 83.1 & 90.9 & 45.9 & 50.3 & 63.1 \\
   	BALMS~\cite{Ren2020Balanced}  & \bf 81.5 & 84.9 & 91.3 & 45.5 & 50.8 & 63.0 \\
   	\midrule
   	\textbf{CDT} \textbf{(Ours)}  & {79.6} & 84.5 & {92.2} & 46.3 & {\bf 51.6} & \bf 65.7 \\
   	\textbf{CDT} \textbf{(Ours) + DRW}  & {81.2} & \bf 85.3 & {\bf 92.4} & \bf 46.4 & {50.9} & 65.1 \\
	\bottomrule
	\end{tabular}
	\label{tab:cifar-table}
\end{table}

\begin{table}[t]
	\small
	\centering
	\tabcolsep 4pt
	\caption{\textbf{Test accuracy (\%) on \emph{step} CIFAR-10 and CIFAR-100.} We show reported results, unless stated otherwise. 
	$\dagger$: our reproduced results. $^\sharp$: best reported results taken from~\cite{cao2019learning,cui2019class}. Two training strategies for CIFAR are explored in the upper~\cite{cao2019learning,cui2019class} and bottom~\cite{Ren2020Balanced} blocks, and we denote the best performance using each strategy in bold. Some methods in~\autoref{tab:cifar-table} do not report step imbalance.}
	\centering
	\begin{tabular}{l@|ccc|ccc}
		\addlinespace
		\toprule
		Dataset           & \multicolumn{3}{c|}{CIFAR-10}                               & \multicolumn{3}{c}{CIFAR-100}                              \\ \midrule
		Imbalance Ratio        & \multicolumn{1}{c}{200} & 100  & 10 & \multicolumn{1}{c}{200} & 100   & 10   \\ 
		\midrule
		ERM~$\dagger$   & 60.0    & 65.3 & 85.1 & 38.7 & 39.9 & 54.6 \\
		Re-sampling~$\dagger$    & 61.3  & 65.0  & 84.5  & 38.0  & 38.4  & 52.1  \\
		Re-weighting~$\dagger$    & 62.6  & 67.3  & 85.8   & 38.2  & 40.1  & 55.7  \\
		Focal~\cite{lin2017focal}~$^\sharp$   & -  & 63.9  & 83.6    & -  & 38.6  & 53.5  \\
		CB~\cite{cui2019class}~$^\sharp$  & - & 61.9  & 84.6   & - & 33.8  & 53.1
		\\
		LDAM~\cite{cao2019learning}    & 60.0$\dagger$  & 66.6  & 85.0    & 39.1$\dagger$  & 39.6  & 56.3  \\
		LDAM + DRW~\cite{cao2019learning}   & {73.6}$\dagger$ & {76.9} & 87.8  & {42.4}$\dagger$ & {45.4} & 59.5  \\
		$\tau$-norm~\cite{kang2019decoupling}~$\dagger$  & 68.8 & 73.0 & 87.3  & {\bf 43.2} & 45.2  & 57.7 \\
		ReMix~\cite{Chou2020Remix} & -  & 69.0  & 86.3   & - & 40.0  & 57.1 \\
		ReMix + DRW~\cite{Chou2020Remix} & -  & 77.8  & 88.3   & - & 46.7  & 60.4 \\
		LogitAdjust~\cite{menon2021longtail} & -  & 72.4  & -   & - & 44.5  & - \\
		\midrule 
		\textbf{CDT} \textbf{(Ours)}   & 70.3 & 76.5 & {88.8}  & 40.0 &  47.0 & {59.6} \\
		\textbf{CDT} \textbf{(Ours) + DRW}  & {\bf 75.5} & \bf 79.1 & {\bf 89.2} & 41.0 & {\bf 47.5} & \bf 60.8 \\
		\midrule
		\midrule
		ERM~$\dagger$~\cite{Ren2020Balanced}  & 63.4 & 71.6 & 89.5 & 40.4 & 41.1 & 60.4 \\
		\midrule
		\textbf{CDT} \textbf{(Ours)}  & {64.9} & 79.7 & {\bf 92.4} & 40.6 & {41.3} & \bf 65.2 \\
		\textbf{CDT} \textbf{(Ours) + DRW}  & {\bf 79.0} & \bf 82.7 & {\bf 92.4} & \bf 44.8 & {\bf 52.0} & 64.3 \\
		\bottomrule
	\end{tabular}
	\label{tab:cifar-table-step}
\end{table}

\subsection{Results on Tiny-ImageNet}
\label{suppl-ssec:tiny}
We show in~\autoref{tab:main-tiny-imagenet-table} and \autoref{tab:tiny-imagenet-table} the results on Tiny-ImageNet with long-tailed and step imbalance ($\rho$ = 10 or 100), respectively.
CDT is on a par with the state of the art.

\begin{table}[t]
	\small
	\caption{{\bf Validation accuracy (\%) on \emph{long-tailed} Tiny-ImageNet.} The best result of each setting (column) is in bold font. $\dagger$: our reproduced results. $^\sharp$: reported results in~\cite{cao2019learning}.}
	\centering
	\begin{tabular}{@{\;}l@{\;}|@{\;}c@{\;}c@{\;}|@{\;}c@{\;}c@{\;}}
	\toprule
	Imbalance Ratio & \multicolumn{2}{c|}{100} & \multicolumn{2}{c}{10} \\ \midrule
	Method          & Top-1       & Top-5       & Top-1       & Top-5      \\ \midrule
	ERM~$\dagger$   & 33.2 & 56.3 & 49.1  & 72.3 \\
	CB~\cite{cui2019class}~$^\sharp$ & 27.3 & 47.4 & 48.4 & 71.1 \\
	LDAM~\cite{cao2019learning}   & 36.0 & 59.5 & 51.9 & 75.2 \\
	LDAM + DRW~\cite{cao2019learning}   & 37.5 & 60.9 & {52.8} & {\bf 76.2} \\
	$\tau$-norm~\cite{kang2019decoupling}~$\dagger$   & 36.4 & 59.8 & 49.6 & 72.8 \\
	\midrule
	\textbf{CDT} \textbf{(Ours)}  & {\bf 37.9} & {\bf 61.4} & 52.7  & 75.6 \\
	\textbf{CDT} \textbf{(Ours)} + DRW  & {37.2} & {60.4} & \bf 53.2  & 75.8 \\
	\bottomrule
	\end{tabular}
	\label{tab:main-tiny-imagenet-table}
\end{table}

\begin{table}[t]
	\small
	\centering
	\caption{{\bf Validation accuracy (\%) on \emph{step} Tiny-ImageNet}. The best result of each setting (column) is in bold font. $\dagger$: our reproduced results. $^\sharp$: reported results in~\cite{cao2019learning}.}
	\label{tab:tiny-imagenet-table}
	\begin{tabular}{l|cc|cc}
		\toprule
		Imbalance Ratio &  \multicolumn{2}{c|}{100} & \multicolumn{2}{c}{10} \\ \midrule
		Method               & Top-1       & Top-5       & Top-1       & Top-5      \\ \midrule
		ERM~$\dagger$   & 36.1 & 54.9 & 48.6 & 72.8 \\
		CB~\cite{cui2019class}~$^\sharp$  & 25.1  & 40.9 & 45.5 & 66.8 \\
		LDAM~\cite{cao2019learning}    & 37.5 & 60.7 & 50.9 & 75.5 \\
		LDAM + DRW~\cite{cao2019learning}   & 39.4 & {61.9} & 52.6 & {\bf 76.7} \\
		$\tau$-norm~\cite{kang2019decoupling}~$\dagger$   & {\bf 40.0} & {61.9} & 51.7 & 75.2 \\
		\midrule
		\textbf{CDT} \textbf{(Ours)}   & {39.6} & 60.9 & {53.3} & 76.2 \\
		\textbf{CDT} \textbf{(Ours)} + DRW  & {39.8} & {\bf 62.5} & \bf 53.5  & 75.2 \\
		\bottomrule
	\end{tabular}
\end{table}

\subsection{Results on iNaturalist}
\label{suppl-ssec:inat}
iNaturalist has 8,142 classes and many of them have scarce instances, making it a quite challenging dataset. 
\autoref{tab:inat-table} shows the results. CDT outperforms many baselines and is on a par with the state of the art. Since large-scale training
is sensitive to batch sizes, the fact that some methods use a batch size larger than 128 (\eg, Remix uses a batch
size of 256) might contribute to the difference.

\begin{table}[t]
	\centering
	\small
	\caption{{\bf Top-1/-5 validation accuracy (\%) on iNaturalist}. See the text for details. $\dagger$: our reproduced results. $^\sharp$: using a batch size larger than 256.}
	\begin{tabular}{l|c|c}
		\toprule
		Method & Top-1 & Top-5 \\
		\midrule
		ERM~\cite{cui2019class}~$\dagger$ & 58.8 &  80.1  \\
		CB~\cite{cui2019class}~$\dagger$ & 61.5 & 80.9 \\
		LDAM~\cite{cao2019learning} & 64.6 & 83.5 \\
		LDAM + DRW~\cite{cao2019learning} & 68.0 &  85.2\\
		BBN~\cite{Zhou2019BBN} & 69.6 & - \\
		$\tau$-norm~\cite{kang2019decoupling}~$^\sharp$ & 69.3 & - \\
		LWS~\cite{kang2019decoupling}~$^\sharp$ & 69.5 & - \\
		Rethinking~\cite{Jamal2020Rethinking} & 67.6 & 86.2 \\
		ReMix~\cite{Chou2020Remix}~$^\sharp$ & 61.3 & 82.3 \\
		ReMix + DRW~\cite{Chou2020Remix}~$^\sharp$ & 70.4 & 87.3 \\
		LogitAdjust~\cite{menon2021longtail}~$^\sharp$ & 68.4  & - \\
		\midrule
		\textbf{CDT} \textbf{(Ours)} & 69.5 & 86.8 \\
		\textbf{CDT} \textbf{(Ours)} + DRW & 69.1 & 86.7 \\
		\bottomrule
	\end{tabular}
	\label{tab:inat-table}
\end{table}

\subsection{Ablation studies}
\begin{figure}[t]
    \centering
    \minipage{0.95\linewidth}
        \centering
        \includegraphics[width=\linewidth]{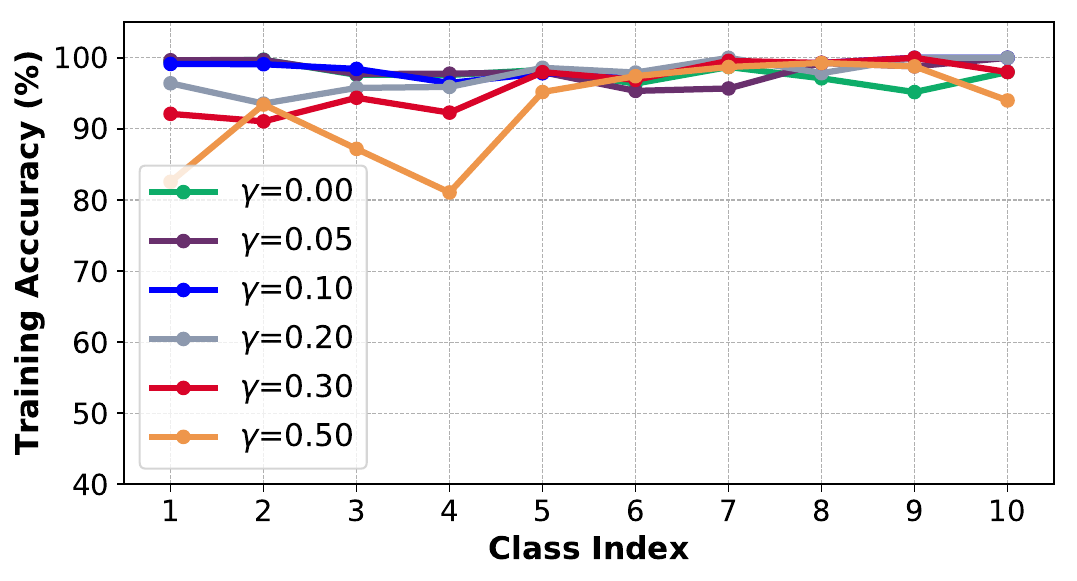}
                \mbox{Training set accuracy}
    \endminipage
    
    \minipage{0.95\linewidth}
        \centering
        \includegraphics[width=\linewidth]{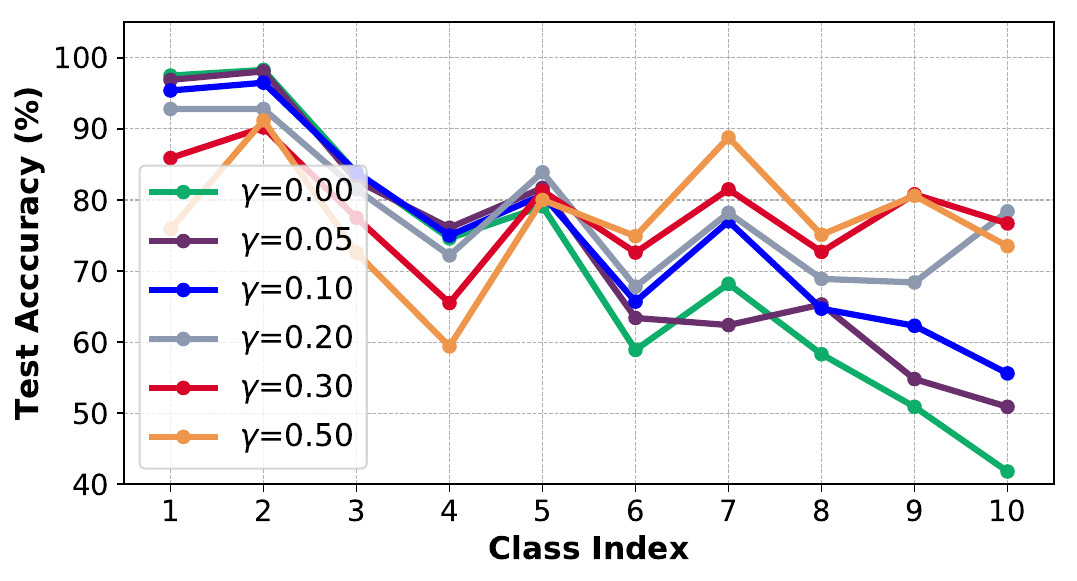}
                \mbox{Test set accuracy}
    \endminipage
    \caption{Effects of $\gamma$ (in computing $a_c$ for CDT) on long-tailed CIFAR-10 ($\rho = 100$) training/test set accuracy. The average test accuracy is $73.2, 75.7, 78.5, \textbf{79.4}, 77.2$ for $\gamma= 0.00, 0.05, 0.10, 0.20, \textbf{0.30}, 0.50$.
    }
    \label{fig:gamma}
\end{figure}

We show the effect of $\gamma$ in computing $a_c$ for our CDT approach in~\autoref{fig:gamma}. We see that increasing $\gamma$ results in better test accuracy on minor classes, but will gradually decrease the accuracy on major classes (in training and testing).

\subsection{Different training objectives and classifiers}
\label{sec:comp}
We provide a comprehensive study on using different training objectives --- ERM, re-sampling, re-weighting, and class-dependent temperatures (CDT) --- in combination with either linear classifiers or nearest center (class mean) classifiers (NCM)~\cite{kang2019decoupling} in \autoref{tab:post_process} (cf. \autoref{ssec:scaling_up} and \autoref{tab:NCM}). Linear classifiers mean that we directly apply the learned ConvNet classifier. 
With linear classifiers, re-weighting outperforms ERM for some cases, while CDT outperforms the others by a notable margin.
Applying the nearest center classifier (NCM) mostly outperforms applying the linear classifiers, except for CDT. In other words, CDT does facilitate learning a ConvNet classifier end-to-end with class-imbalanced data.

\begin{table}[t]
	\centering
	\small
	\tabcolsep 3pt
	\caption{\small\textbf{Test accuracy on long-tailed CIFAR-10/-100 with different training objectives and classifiers.} The highest accuracy in each setting (column) with each classification method (row block) is in bold font. ``LT'' denotes ``long-tailed''.}
	\begin{tabular}{@{\;}c@{\;}|@{\;}ccc@{\;}|@{\;}ccc@{\;}}
		\addlinespace
		\toprule
		\multicolumn{ 1}{c|}{Data Type} & \multicolumn{ 3}{|c|@{\;}}{LT-CIFAR-10} & \multicolumn{ 3}{c}{LT-CIFAR-100} \\
		\midrule
		\multicolumn{ 1}{c|}{Imbalance Ratio} & 200   & 100   & 10    & 200   & 100   & 10 \\
		\midrule
		\multicolumn{ 7}{c}{Linear} \\
		\midrule
		ERM   & 65.6 & 71.1 & 87.2 & 35.9 & 40.1 & 56.9 \\
		Re-sampling & 64.4 & 71.2 & 86.5 & 30.6 & 34.7 & 54.2 \\
		Re-weighting & 68.6 & 72.6 & 87.1  & 35.0 & 40.5 & 57.3 \\
		CDT   & {\bf 74.7} & {\bf 79.4} & {\bf 89.4} & {\bf 40.5} & {\bf 44.3} & {\bf 58.9} \\
		\midrule
		\midrule
		\multicolumn{7}{c}{NCM} \\\midrule
		ERM   & 73.2 & 77.0 & {\bf 88.2} & {\bf 39.4} & {\bf 42.8} & {\bf 55.8} \\
		Re-sampling & 66.0 & 72.3 & 87.3 & 32.0 & 36.1  & 54.1 \\
		Re-weighting & 66.3 & 77.0 & 88.0 & 37.3 & 42.5 & 55.8 \\
		CDT   & {\bf 74.8} & {\bf 78.8} & 88.1 & 37.9 & 42.7 & 55.7 \\
		\bottomrule
	\end{tabular}
	\label{tab:post_process}
\end{table}

\section{Experimental Analyses}\label{sec:exp_analysis}
We provide further analyses on ERM and CDT. We mainly focus on the long-tailed CIFAR-10.

\begin{figure*}[t!]
    \centering
    \minipage{0.24\textwidth}
        \centering
        \mbox{Training set accuracy}
        \includegraphics[width=1.\linewidth]{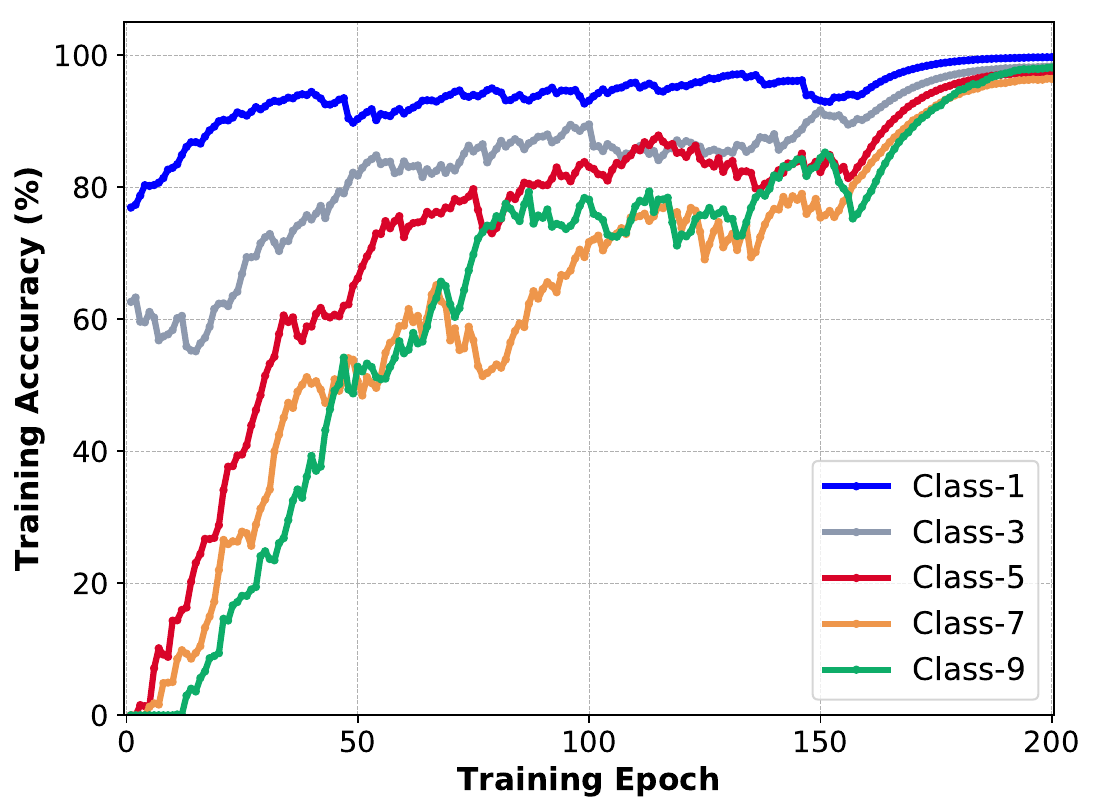}
    \endminipage\hspace{0.05cm}
    \minipage{0.24\textwidth}
        \centering
        \mbox{Test set accuracy}
        \includegraphics[width=1.\linewidth]{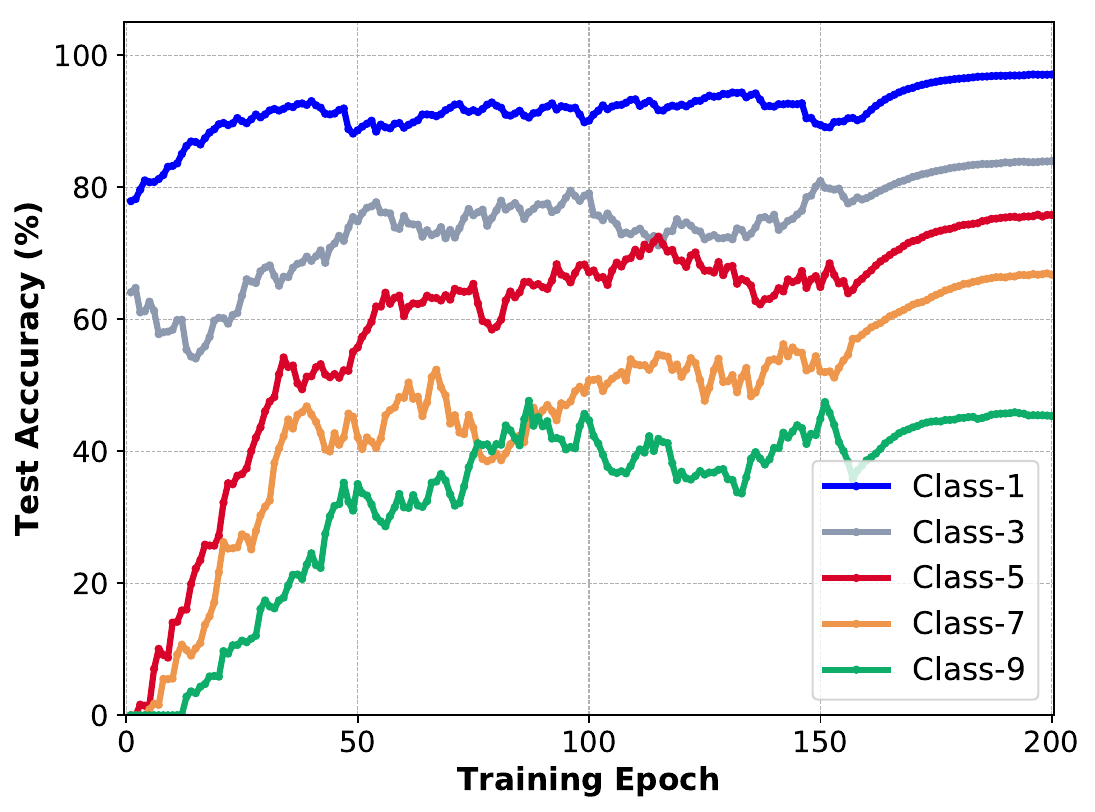}
        \endminipage\hspace{0.05cm}
    \minipage{0.24\textwidth}
        \centering
        \mbox{Classifier norm}
        \includegraphics[width=1.\linewidth]{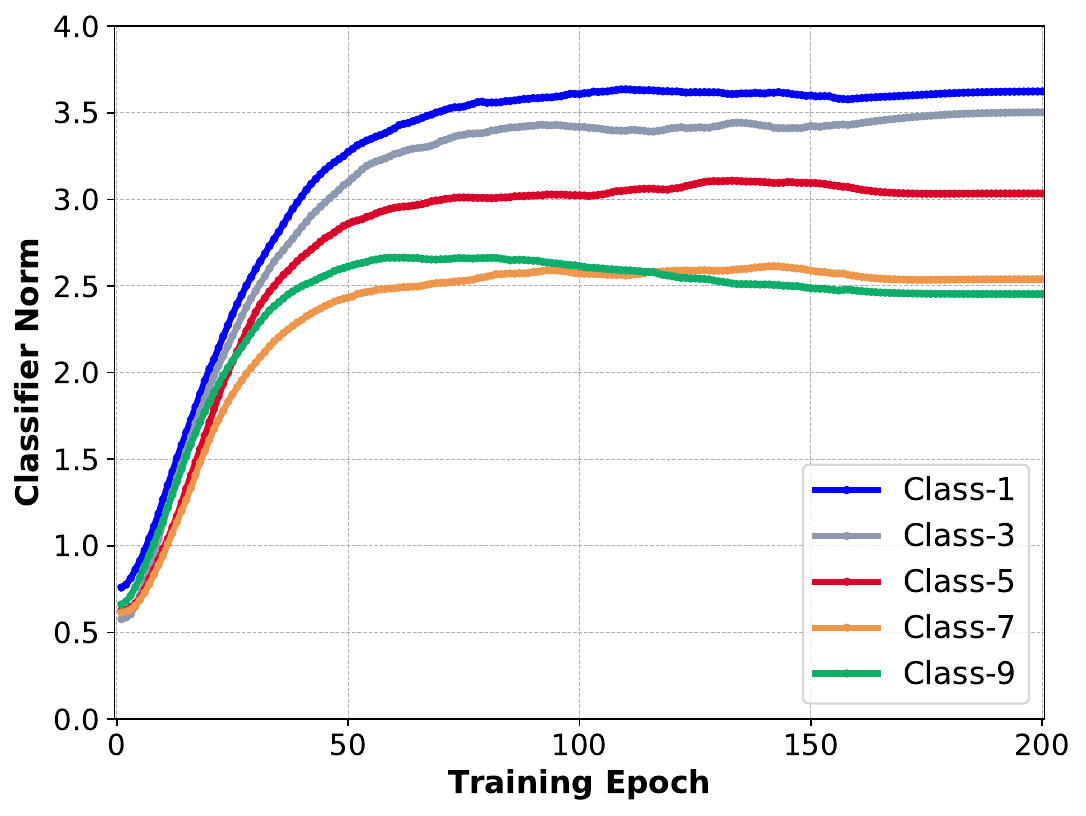}
    \endminipage\hspace{0.05cm}
    \minipage{0.24\textwidth}
        \centering
        \mbox{Feature deviation}
        \includegraphics[width=1.\linewidth]{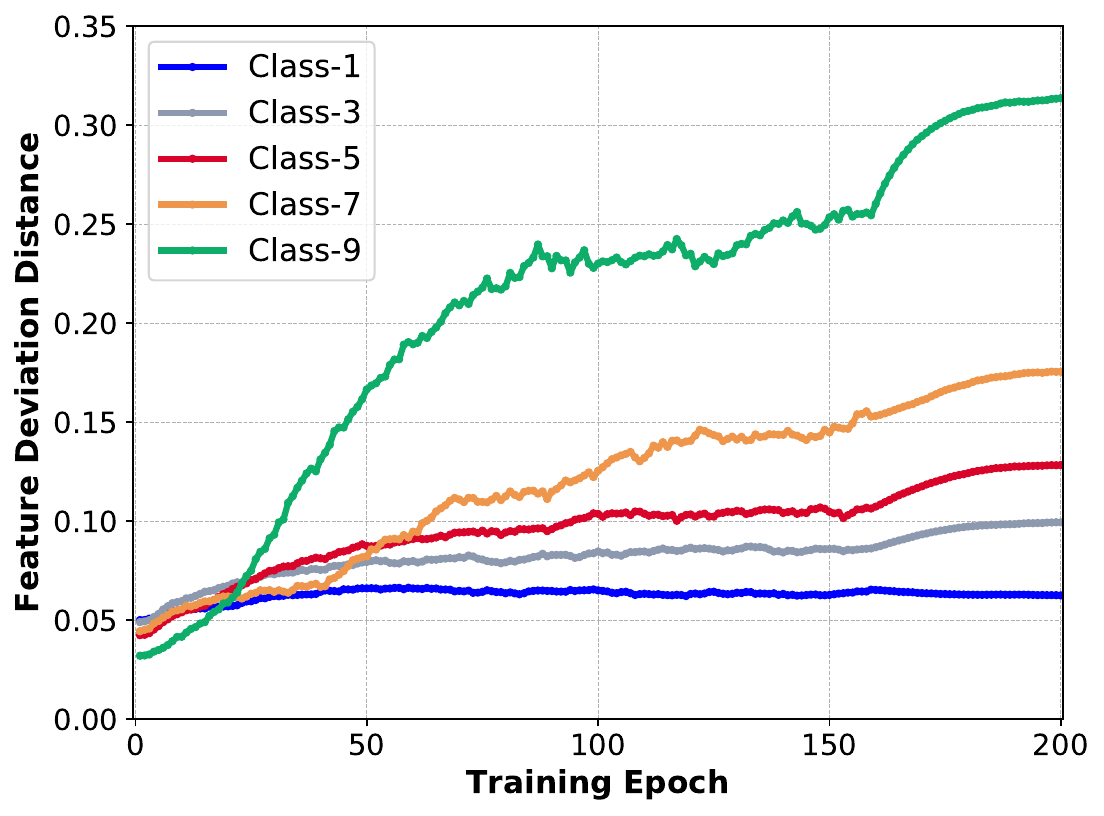}
    \endminipage\hspace{0.05cm}
        \minipage{0.24\textwidth}
        \includegraphics[width=1.\linewidth]{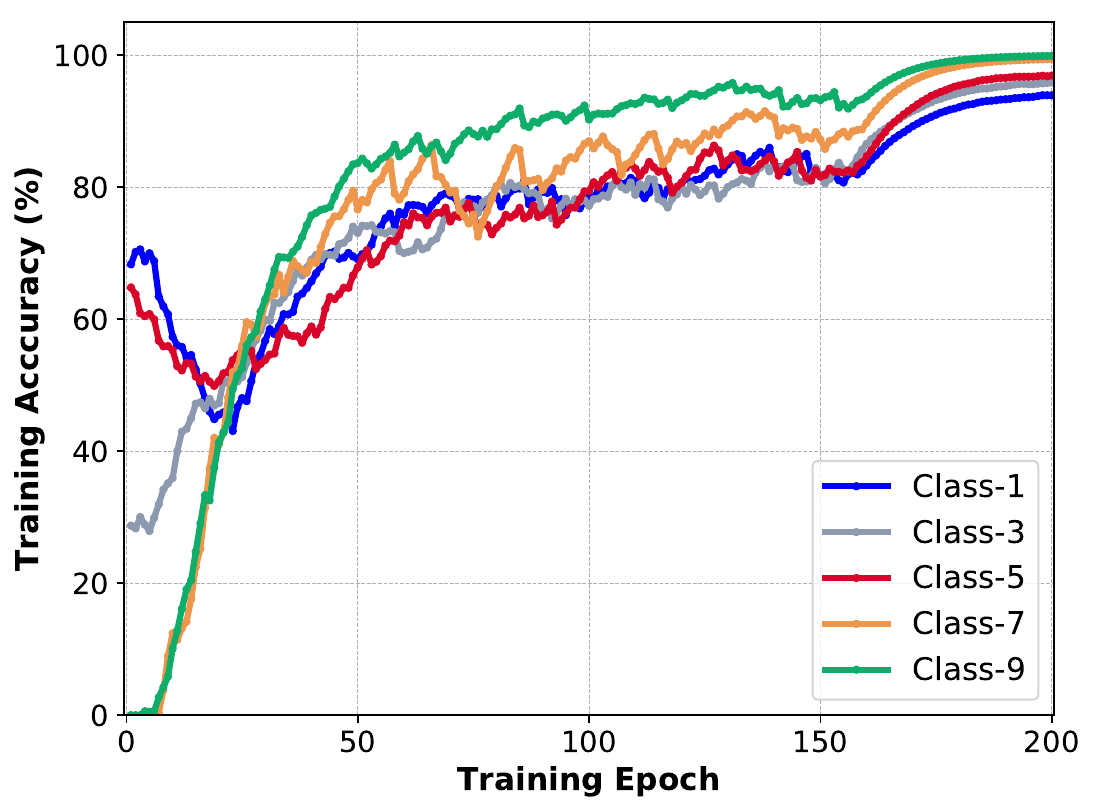}
    \endminipage\hspace{0.05cm}
    \minipage{0.24\textwidth}
        \includegraphics[width=1.\linewidth]{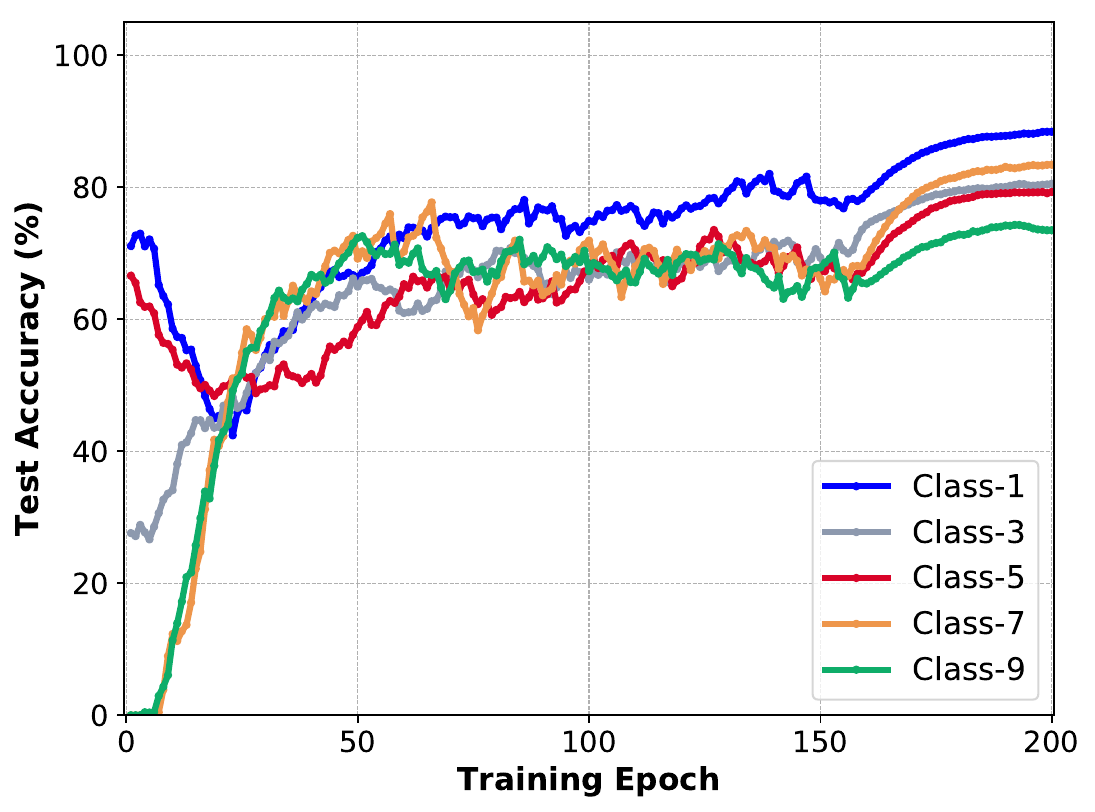}
        \endminipage \hspace{0.05cm}
    \minipage{0.24\textwidth}
        \includegraphics[width=1.\linewidth]{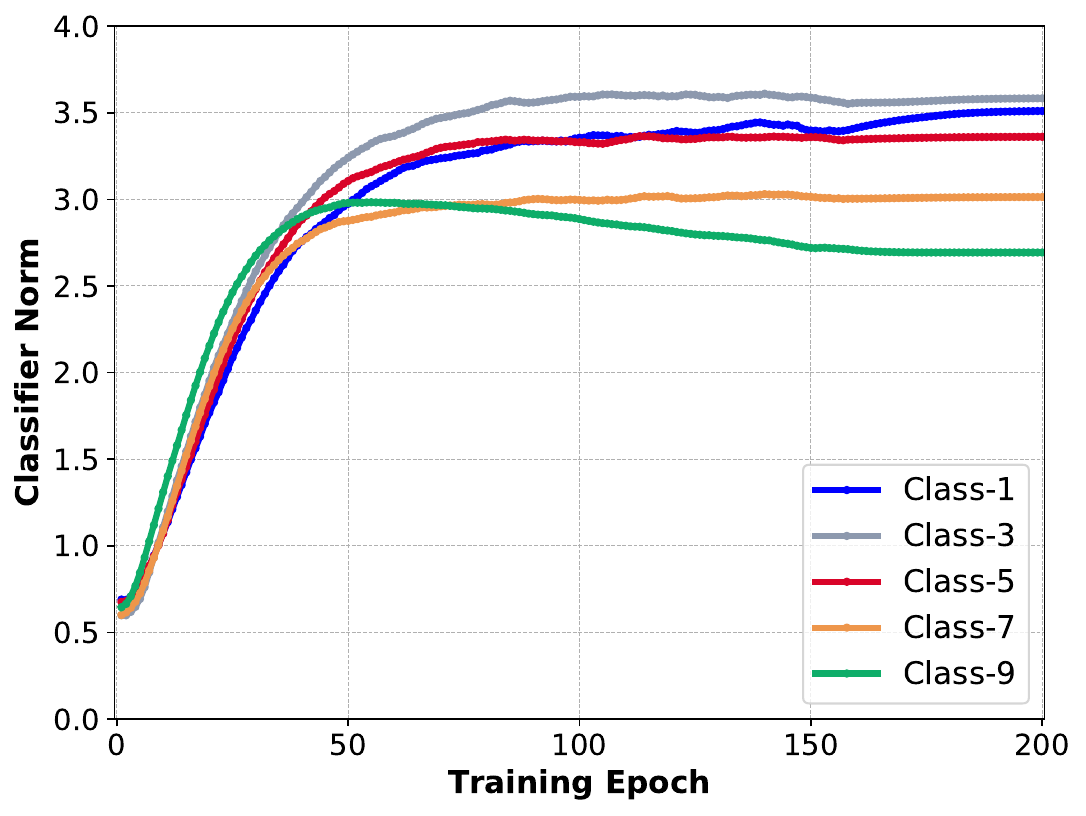}
    \endminipage\hspace{0.05cm}
    \minipage{0.24\textwidth}
        \includegraphics[width=1.\linewidth]{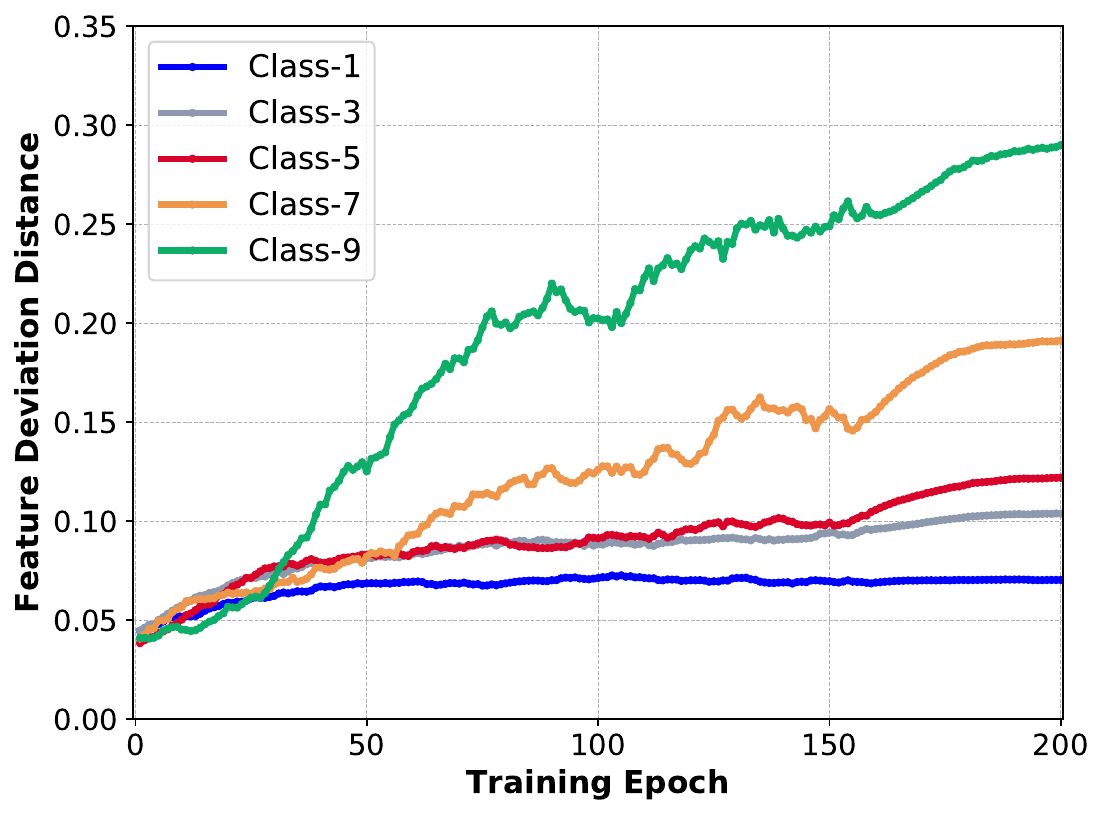}
    \endminipage\hspace{0.05cm}
    \caption{\textbf{Statistics of ERM (\textbf{top row}) and CDT with $\gamma = 0.3$ (\textbf{bottom row}) on long-tailed CIFAR-10 ($\rho = 100$).} We show for each epoch the training set accuracy, test set accuracy, classifier norms $\|\vw_c\|_2$, and feature deviation (Euclidean distance between training and test feature means per class; \ie, \autoref{eq_feat_dev}). We show five classes for brevity and apply the built-in smoothing of TensorBoard for better visualization. Class $c = 1,3,5,7,9$ has $5000, 1796, 645, 232, 83$ training instances, respectively. The features are $\ell_2$-normalized first.}
    \label{fig:main-erm_stat}
\end{figure*}

\subsection{Training and testing along epochs}
We plot the classifier's performance (trained with either ERM or our CDT) and its statistics (\ie, classifier norms and feature deviation) along the training epochs in~\autoref{fig:main-erm_stat}. We see that using ERM, the training accuracy of the major classes immediately goes to nearly $100\%$ (using \autoref{eq:test}) in the early epochs. The classifier norms of all the classes increase sharply in the early epochs, and those of major classes are larger than those of minor classes. The feature deviation between training and test means also increases, but in a reverse order (minor classes have higher deviation). \emph{Our CDT does not eliminate the biased trends of classifier norms and feature deviation, but makes the training set and test set accuracy more balanced across classes.} As training with balanced data will likely have balanced statistics over all the classes, we hypothesize that balancing these statistics in training would facilitate class-imbalanced deep learning.

\begin{figure*}[t]
	\centering
	\minipage{0.24\textwidth}
	\centering
	\includegraphics[width=1.\linewidth]{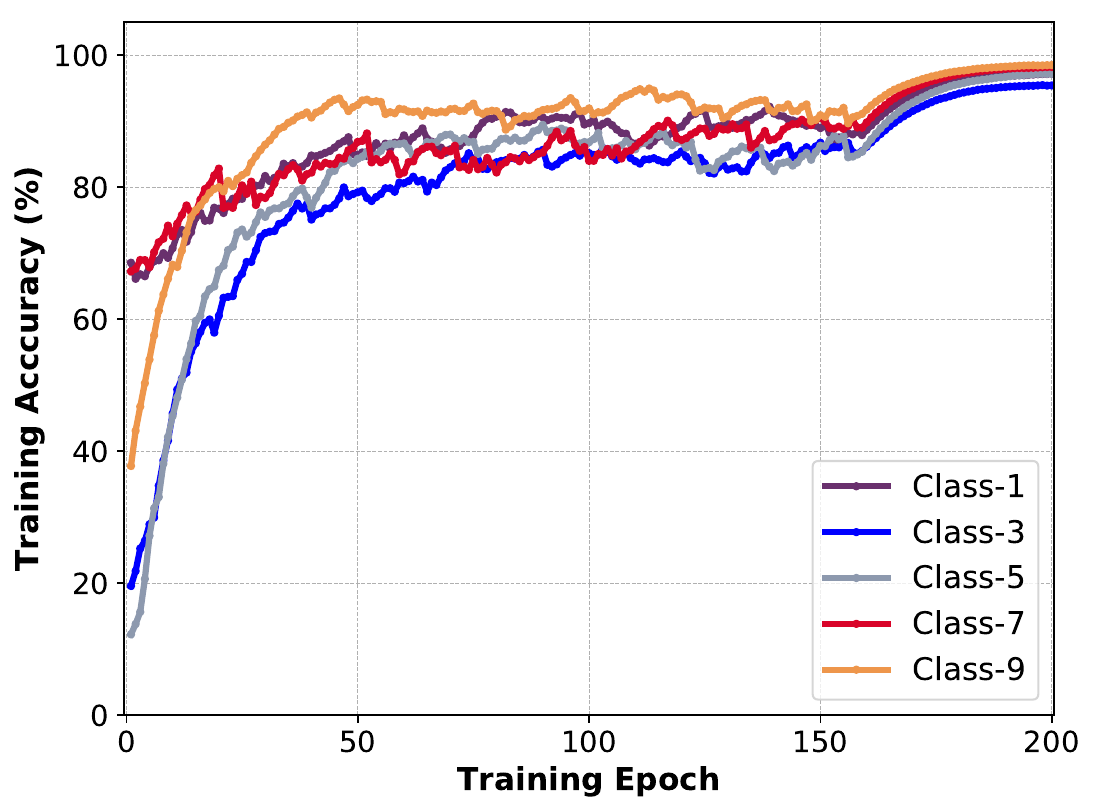}
		\mbox{Training set accuracy}
	\endminipage\hspace{0.05cm}
	\minipage{0.24\textwidth}
	\centering
	\includegraphics[width=1.\linewidth]{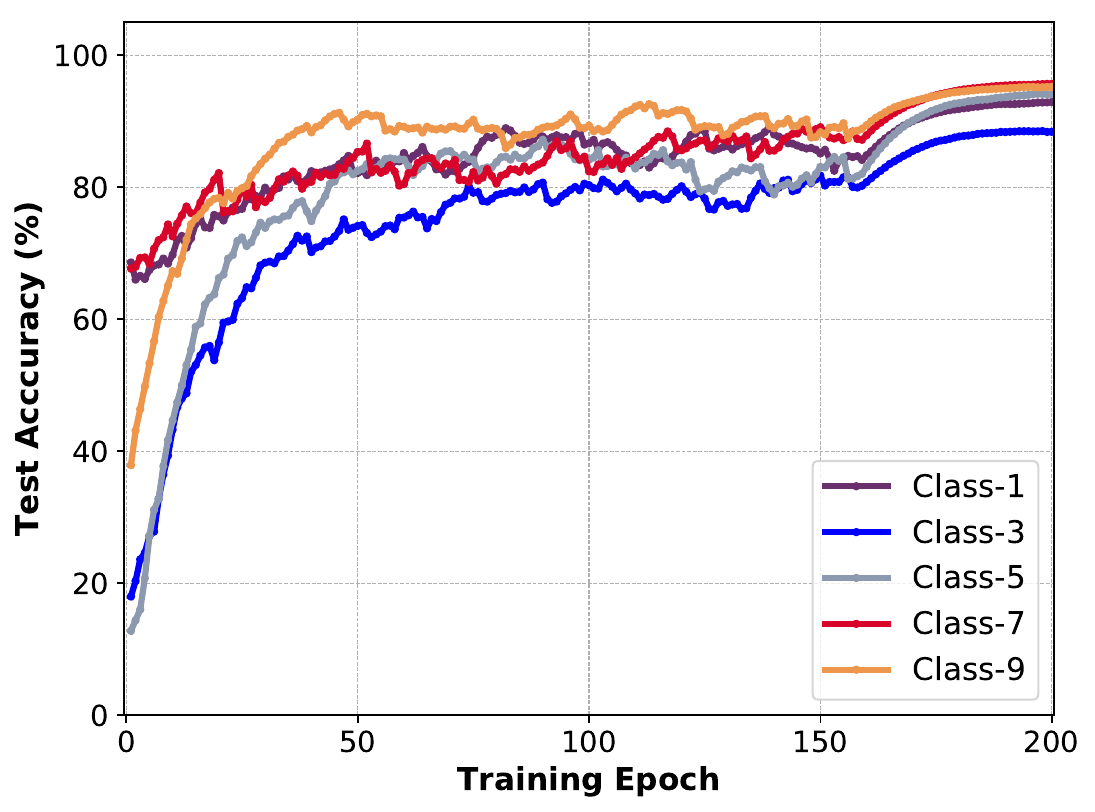}
		\mbox{Test set accuracy}
	\endminipage\hspace{0.05cm}
	\minipage{0.24\textwidth}
	\centering
	\includegraphics[width=1.\linewidth]{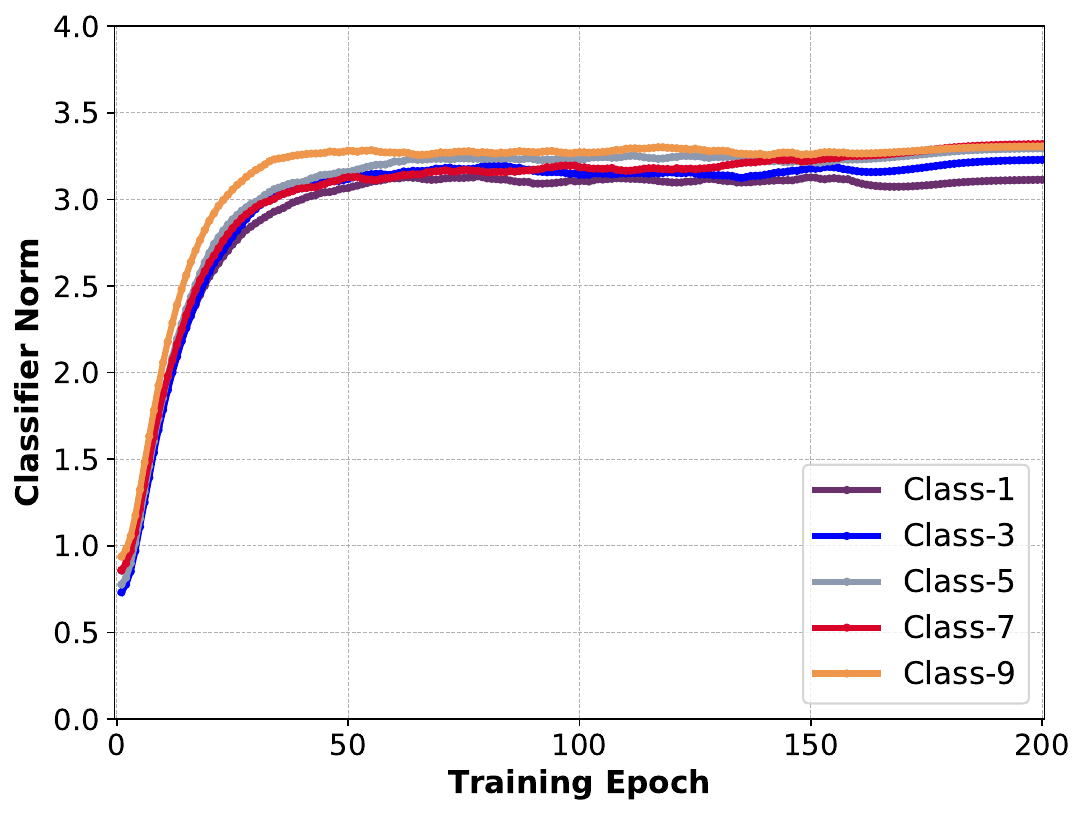}
		\mbox{Classifier norm}
	\endminipage\hspace{0.05cm}
	\minipage{0.24\textwidth}
	\centering
	\includegraphics[width=1.\linewidth]{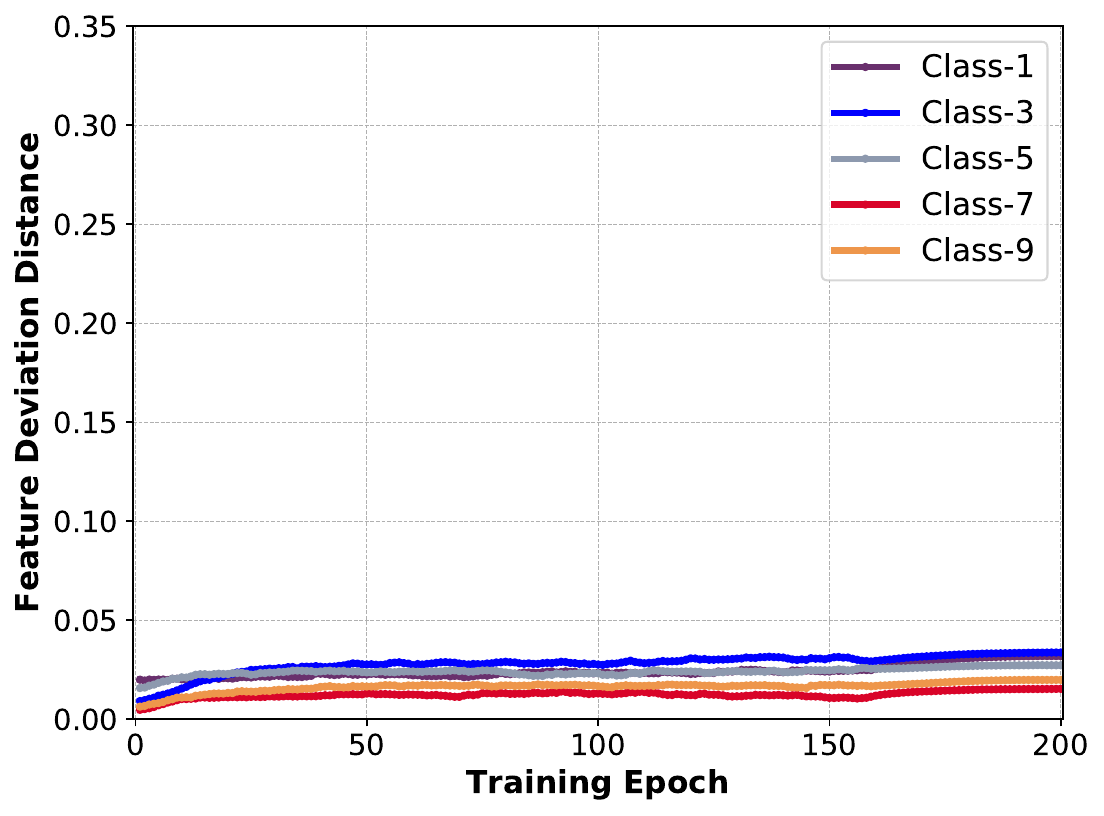}
		\mbox{Feature deviation}
	\endminipage\hspace{0.05cm}
	\caption{\textbf{Statistics of ERM on original balanced CIFAR-10.}  We show for each epoch the training set accuracy, test set accuracy, classifier norms $\|\vw_c\|_2$, and feature deviation (Euclidean distance between training and test feature means per class; \ie, \autoref{eq_feat_dev}). We show five classes for brevity and apply the built-in smoothing of TensorBoard for better visualization. The features have been $\ell_2$-normalized first.}
	\label{fig:erm_stat}
\end{figure*}

To verify this, we train a ConvNet classifier using ERM on the original balanced CIFAR-10. We show the results in~\autoref{fig:erm_stat}: we see similar trends of statistics among classes.

\subsection{Features vs. classifiers}
Learning with class-imbalanced data affects both the linear classifiers and the ConvNet features, but which of them is more unfavorable? To answer this, we collect a balanced training set from CIFAR-10 which contains the same number of total training instances as long-tailed CIFAR-10 ($\rho=100$). We then investigate:
\begin{itemize}
    \item Train a ConvNet with the balanced (imbalanced) set.
    \item Train a ConvNet with the balanced (imbalanced) set, freeze the feature extractor $f_{\vtheta}(\cdot)$ but remove the linear classifiers $\{\vw_c\}_{c=1}^C$, and retrain only the linear classifiers using the imbalanced (balanced) set.
\end{itemize}
For this experiment, we apply ERM for training.

\autoref{tab:final_analysis} shows the results. We found that the negative influence of class-imbalanced data on ConvNet features is severer than on linear classifiers: imbalanced features with balanced classifiers perform worse than balanced features with imbalanced classifiers. These results suggest that feature deviation (trained with imbalanced data) is a fundamental issue to be resolved in class-imbalanced deep learning.

\begin{table}[t]
	\small
	\centering
	\caption{Effects of imbalanced data on classifiers and features. We collect a balanced CIFAR-10 whose number of training images is the the same as long-tailed CIFAR-10 ($\rho=100$).}
	\begin{tabular}{c|c|c}
		\toprule
		Feature / classifier & balanced & imbalanced \\ \hline
		balanced & 85.6 & 82.2	\\ 
		imbalanced & 77.7 & 71.1 \\
		\bottomrule
	\end{tabular}
	\label{tab:final_analysis}
\end{table}

\begin{figure*}[t]
	\centering
	\minipage{0.45\textwidth}
	\centering
	\includegraphics[width=1.\linewidth]{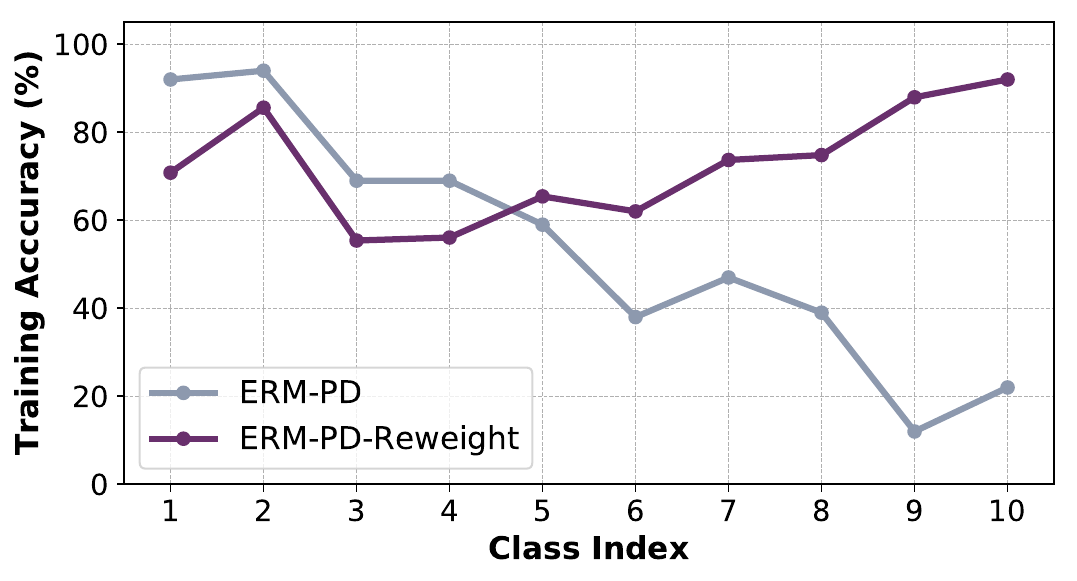}
	\mbox{\textbf{Long-tailed CIFAR-10:} training set accuracy}
	\endminipage\hspace{0.05cm}
	\minipage{0.45\textwidth}
	\centering
	\includegraphics[width=1.\linewidth]{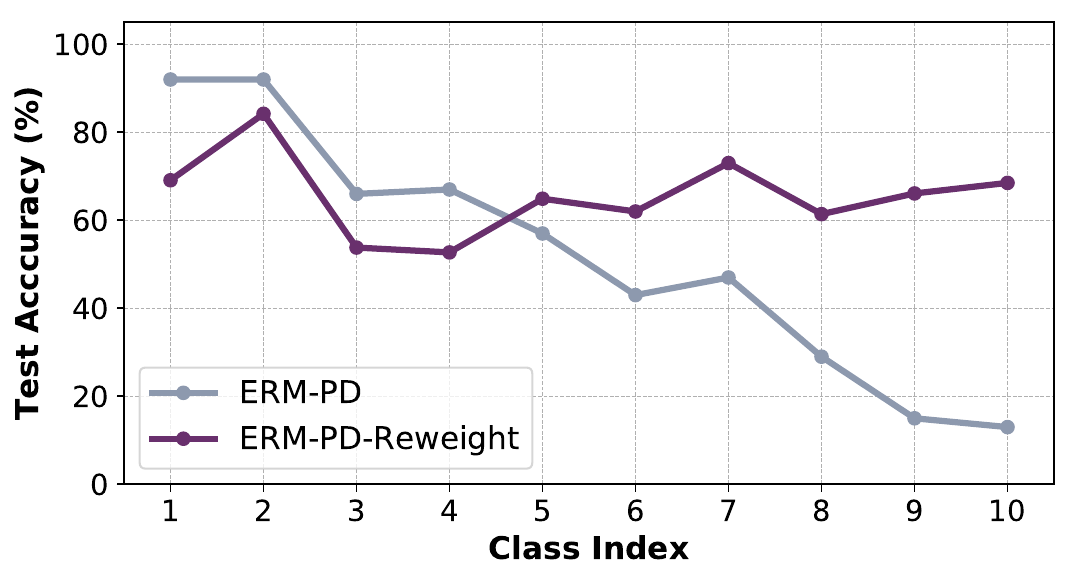}
	\mbox{\textbf{Long-tailed CIFAR-10:} test set accuracy}
	\endminipage\hspace{0.05cm}
	\minipage{0.45\textwidth}
	\centering
	\includegraphics[width=1.\linewidth]{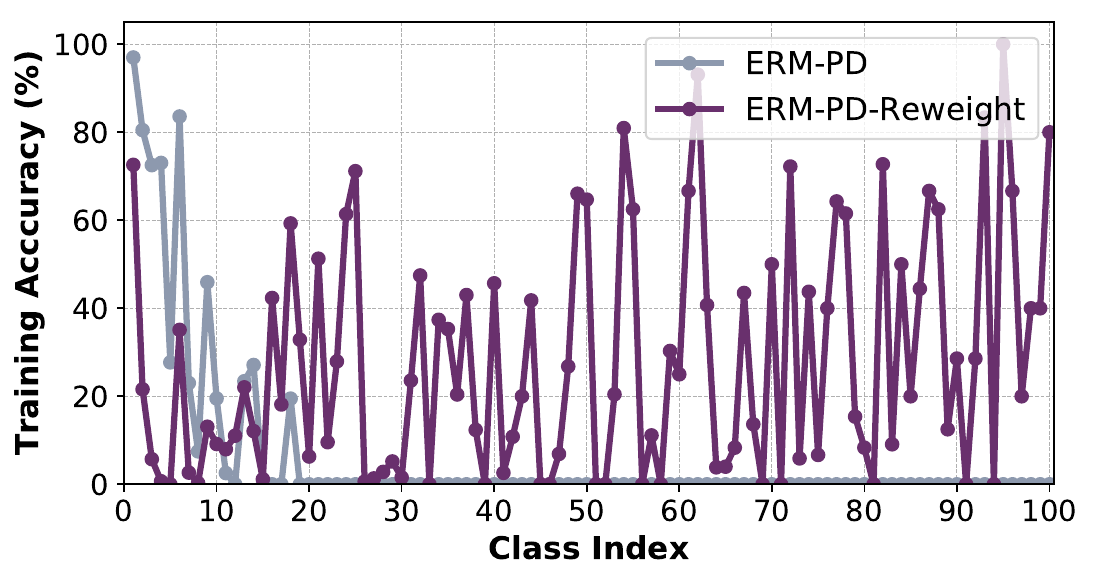}
		\mbox{\textbf{Long-tailed CIFAR-100:} training set accuracy}
	\endminipage\hspace{0.05cm}
	\minipage{0.45\textwidth}
	\centering
	\includegraphics[width=1.\linewidth]{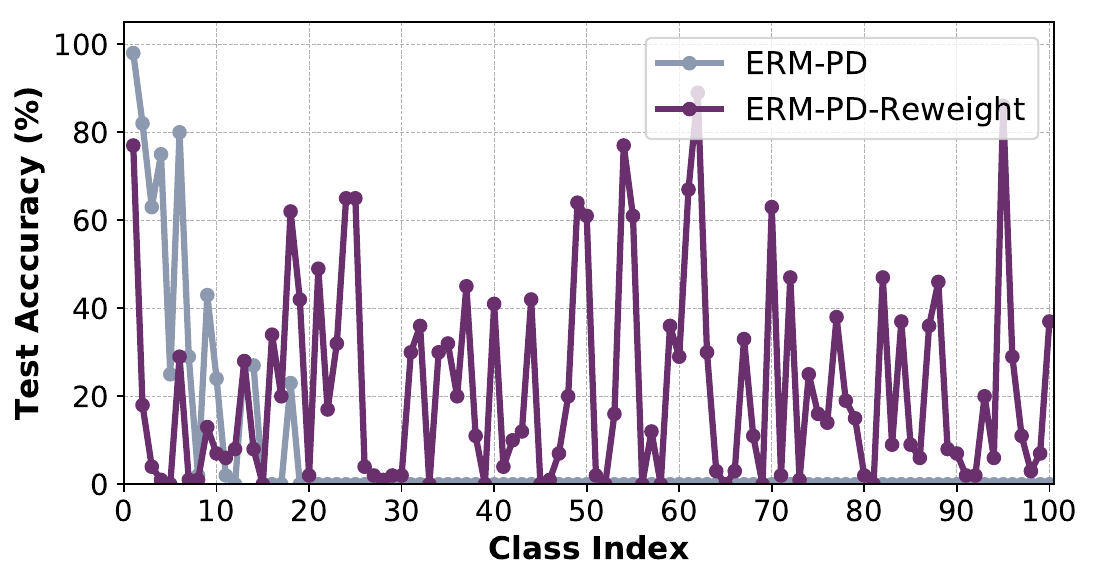}
		\mbox{\textbf{Long-tailed CIFAR-100:} test set accuracy}
	\endminipage\hspace{0.05cm}
	
	\caption{\textbf{The effects of learning with class-imbalanced data using traditional machine learning approaches.}  We train a linear logistic regression classifier using ERM or re-weighting on pre-defined features. We experiment on long-tailed CIFAR-10 and CIFAR-100 ($\rho=100$). We show the training set accuracy and test set accuracy. We see under-fitting to minor classes using ERM, and re-weighting can effectively alleviate the problem.}
	\label{fig:linear}
\end{figure*}

\begin{figure*}[t]
	\centering
	\minipage{0.33\textwidth}
	\centering
	\includegraphics[width=1.\linewidth]{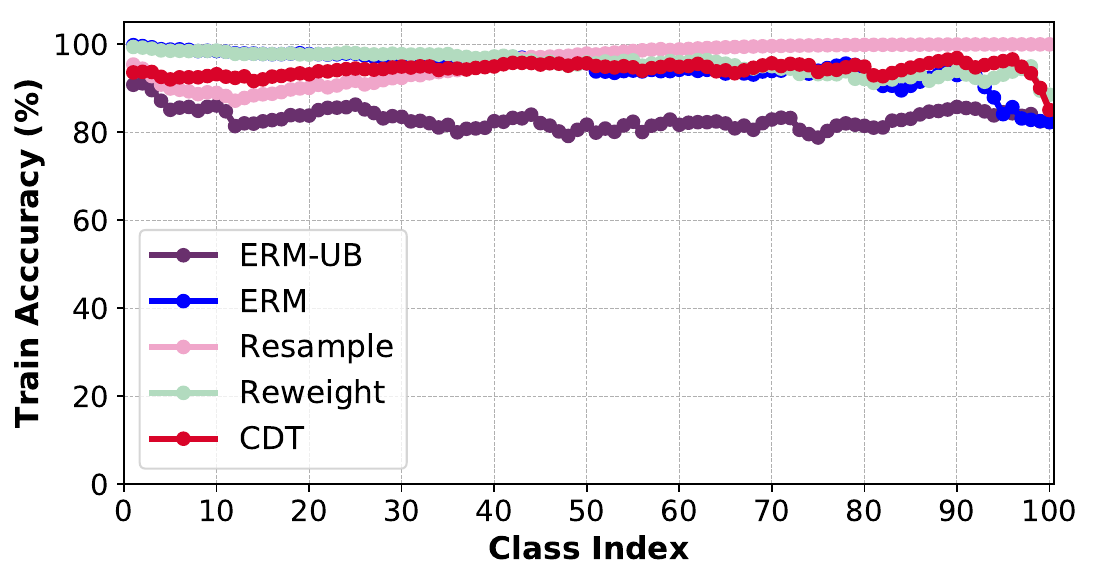}
	\mbox{(a) Training set accuracy}
	\endminipage
	\minipage{0.33\textwidth}
	\centering
	\includegraphics[width=1.\linewidth]{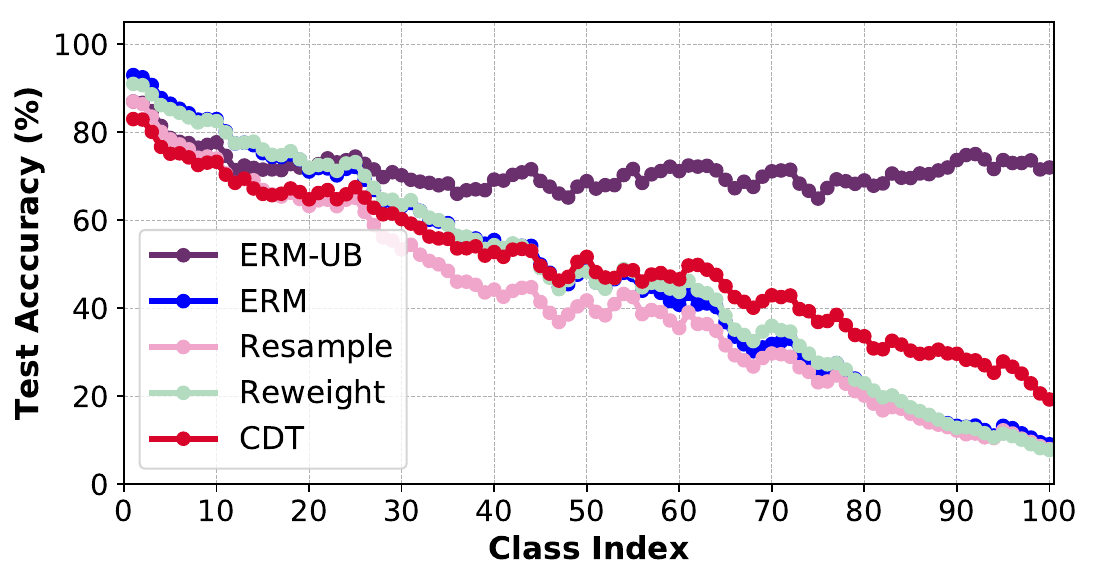}
	\mbox{(b) Test set accuracy}
	\endminipage
	\minipage{0.33\textwidth}
	\centering
	\includegraphics[width=1.\linewidth]{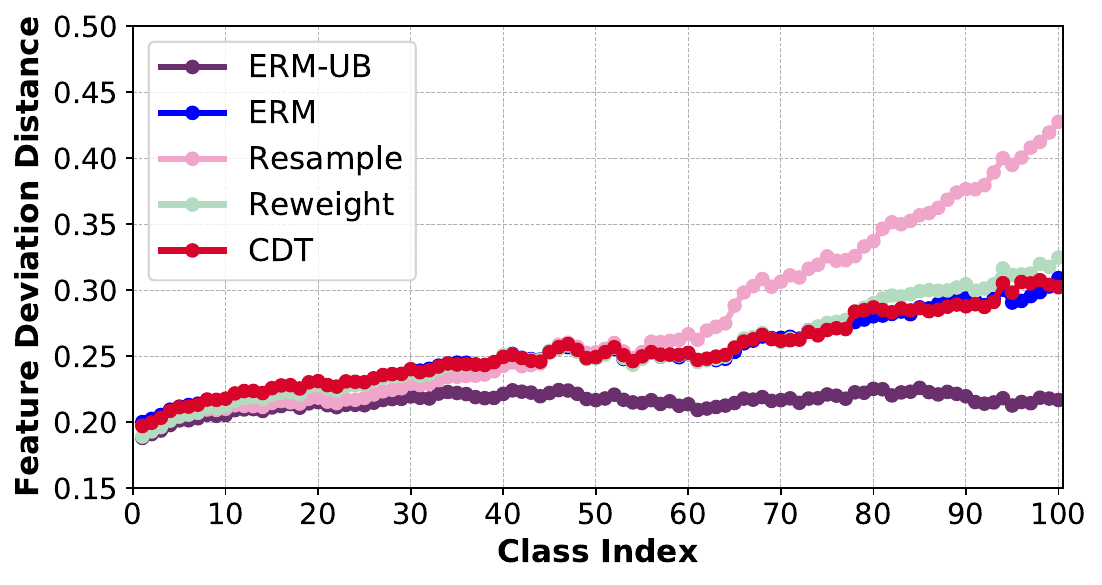}
	\mbox{(c) Feature deviation}
	\endminipage
	\caption{\textbf{The effects of learning with class-imbalanced data on long-tailed CIFAR-100 ($\rho=100$).} We train a ConvNet using ERM, re-weighting, re-sampling, and our \textbf{class-dependent temperatures (CDT)}. We also investigate the upper bound by training a ConvNet classifier using ERM on the original balanced CIFAR-100, denoted as ERM-UB. We show the (a) training set accuracy, (b) test set accuracy, and (c) feature deviation (\autoref{eq_feat_dev}) for each class. We see over-fitting to minor classes for ConvNet classifiers trained with class-imbalanced data. CDT can \emph{compensate for the effect of feature deviation} (but not directly reduce deviation) and thus leads to a higher test set accuracy.}
	\label{fig:fig2}
\end{figure*}

\begin{figure*}[t]
	\centering
	\minipage{0.33\textwidth}
	\centering
	\includegraphics[width=1.\linewidth]{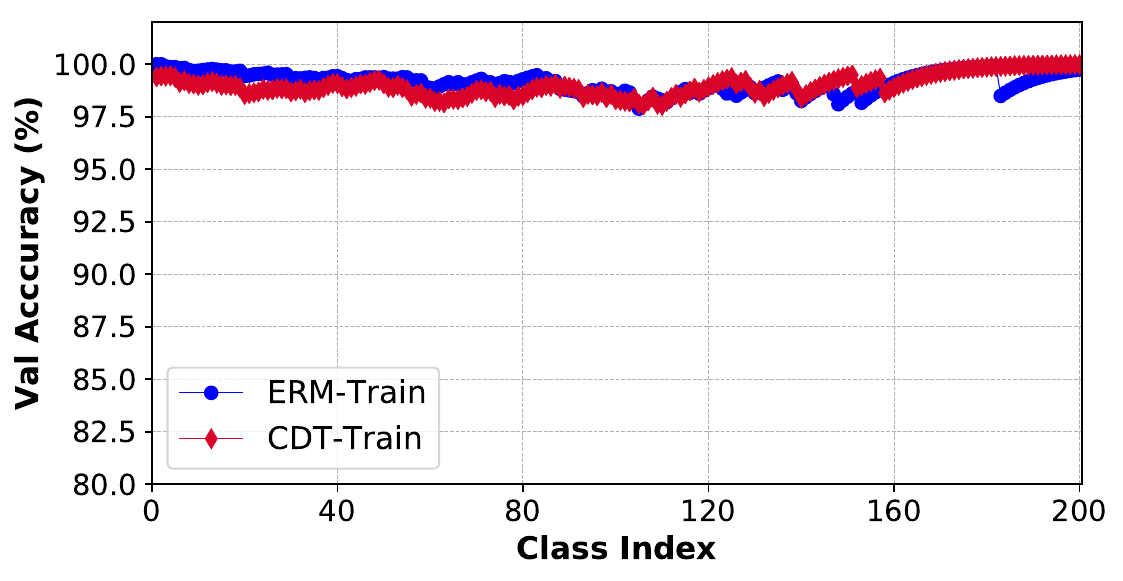}
		\mbox{(a) Training set accuracy}
	\endminipage
	\minipage{0.33\textwidth}
	\centering
	\includegraphics[width=1.\linewidth]{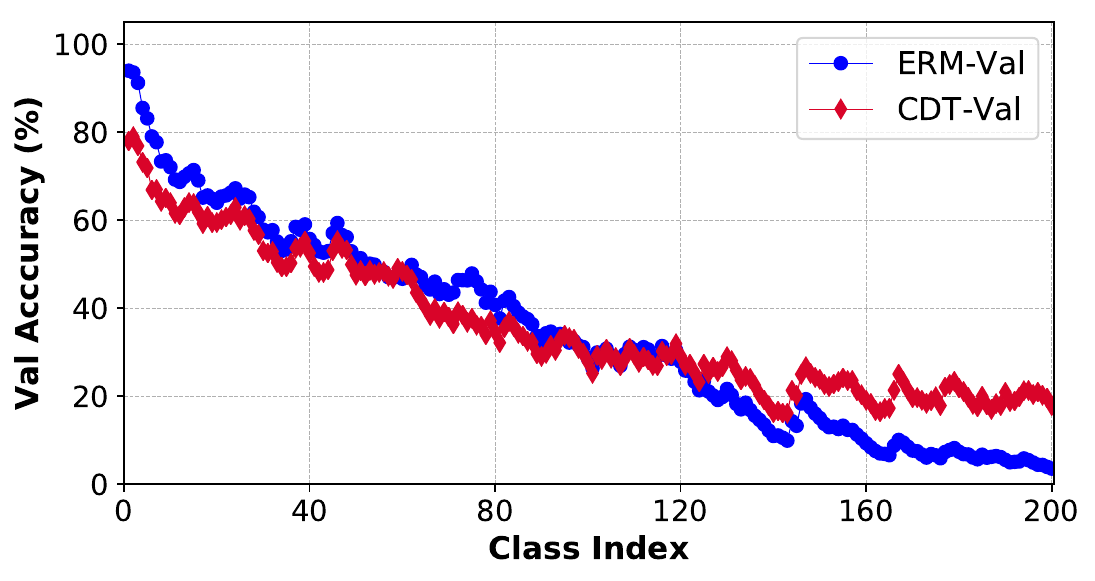}
		\mbox{(b) Validation set accuracy}
	\endminipage
	\minipage{0.33\textwidth}
	\centering
	\includegraphics[width=1.\linewidth]{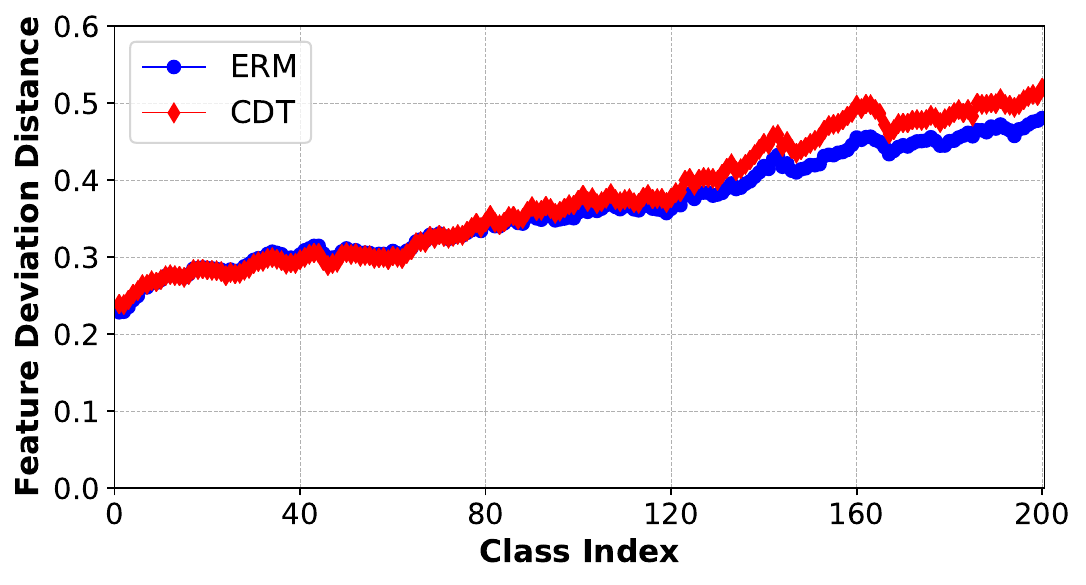}
	\mbox{(c) Feature deviation}
	\endminipage
	\caption{\textbf{The effects of learning with class-imbalanced data on long-tailed Tiny-ImageNet ($\rho=100$).} We train a ConvNet using ERM and our \textbf{class-dependent temperatures (CDT)}. We show the (a) training set accuracy, (b) validation set accuracy, and (c) feature deviation (\autoref{eq_feat_dev}) for each class. CDT can \emph{compensate for the effect of feature deviation} (but not directly reduce deviation) and thus leads to a higher test set accuracy.}
	\label{sup-fig:tiny}
\end{figure*}

\begin{figure*}[t]
	\centering
	\minipage{0.33\textwidth}
	\centering
	\includegraphics[width=1.\linewidth]{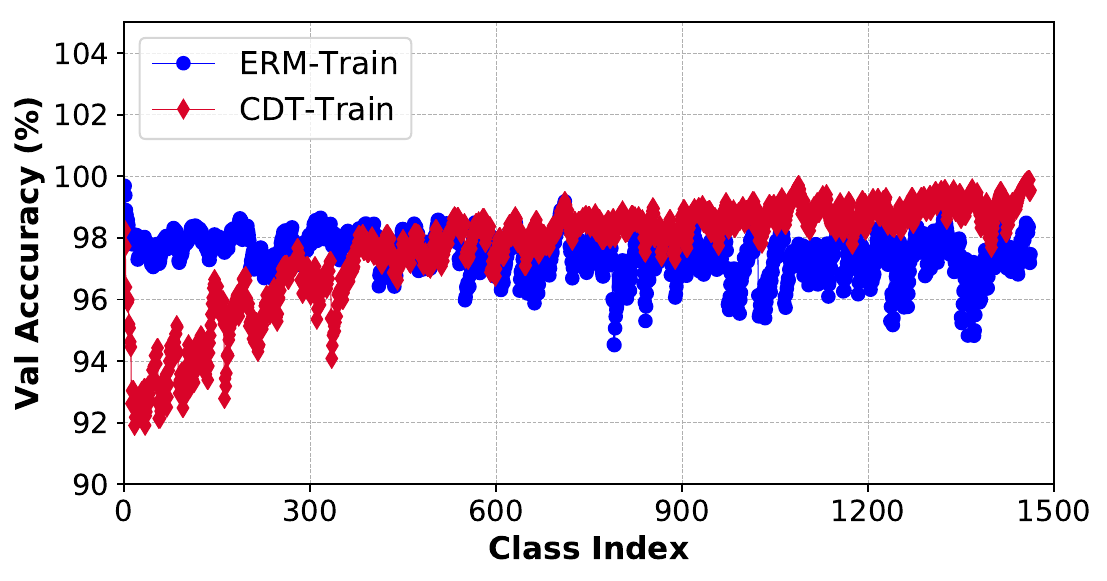}
		\mbox{(a) Training set accuracy}
	\endminipage
	\minipage{0.33\textwidth}
	\centering
	\includegraphics[width=1.\linewidth]{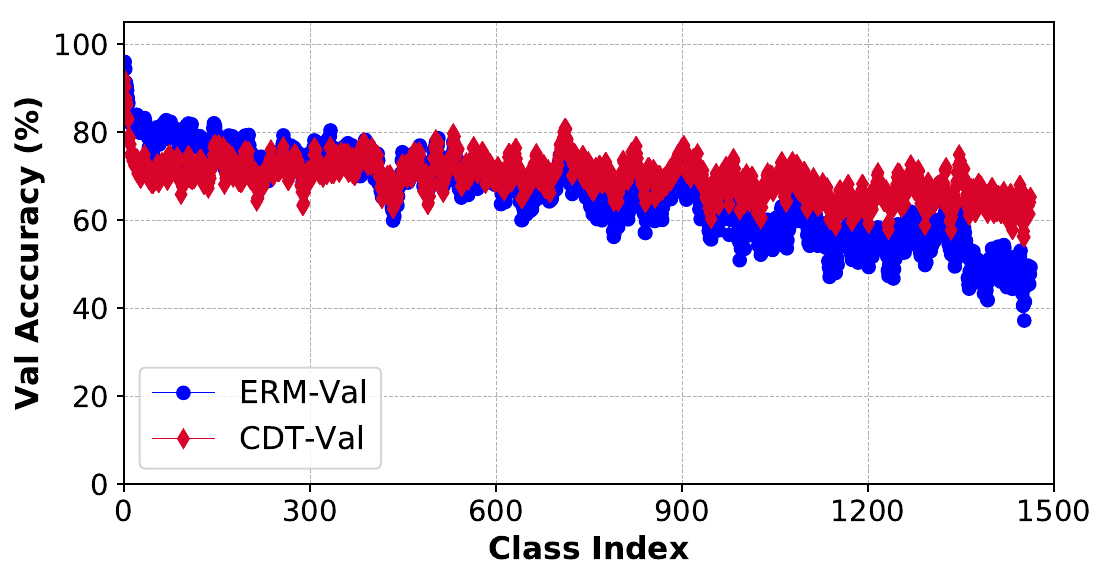}
		\mbox{(b) Validation set accuracy}
	\endminipage
	\minipage{0.33\textwidth}
	\centering
	\includegraphics[width=1.\linewidth]{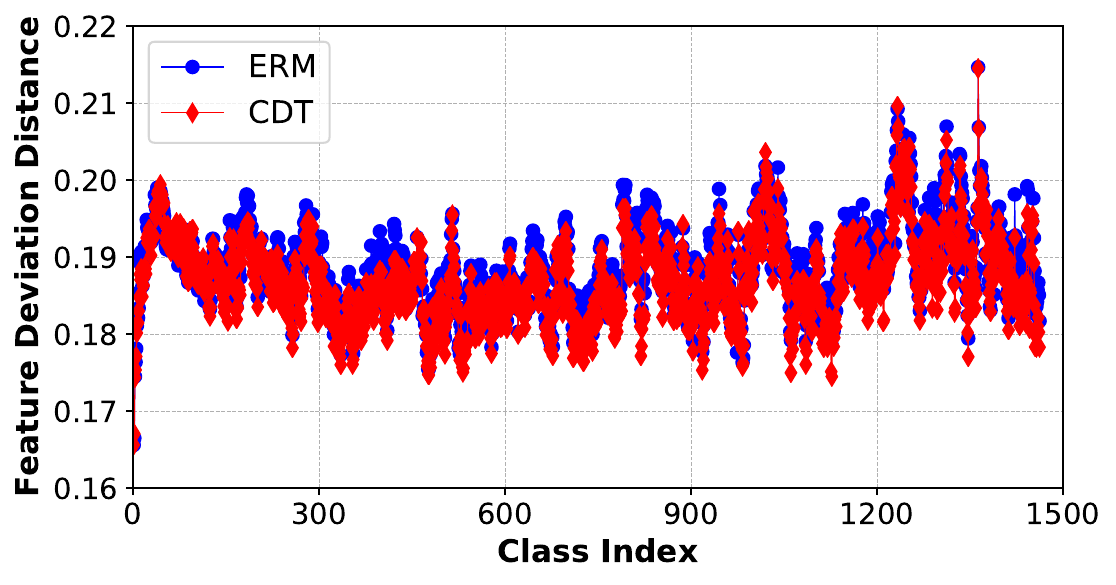}
		\mbox{(c) Feature deviation}
	\endminipage
	\caption{\textbf{The effects of learning with class-imbalanced data on iNaturalist.} We experiment on a re-split set with $1,462$ classes. (See~\autoref{suppl-ssec:class-imbalanced-all}.) We train a ConvNet using ERM and our \textbf{class-dependent temperatures (CDT)}. We show the (a) training set accuracy, (b) validation set accuracy, and (c) feature deviation (\autoref{eq_feat_dev}) for each class. CDT can \emph{compensate for the effect of feature deviation} (but not directly reduce deviation) and thus leads to a higher test set accuracy.}
	\label{sup-fig:inat}
\end{figure*}

\subsection{Results using traditional machine learning}
\label{ssec:linear}
We investigate learning linear logistic regression classifiers using pre-defined features on long-tailed CIFAR. Specifically, we pre-train a ConvNet classifier using ResNet-32~\cite{he2016deep} on the original balanced Tiny-ImageNet (see also~\autoref{ssec:traditional_deep} of the main paper). We then use the learned feature extractor to extract features for CIFAR images. We then learn a linear logistic regression classifier using ERM or using ERM with re-weighting. \autoref{fig:linear} shows the results. On both long-tailed CIFAR-10 and CIFAR-100, we see under-fitting to minor classes using ERM. We also see that re-weighting can effectively alleviate the problem, leading to better overall accuracy. These results further suggest that, popular techniques for traditional machine learning approaches may not work well for deep learning approaches in class-imbalanced learning, and vice versa.

\subsection{Effects of learning with class-imbalanced data}
\label{suppl-ssec:class-imbalanced-all}

We follow \autoref{fig:2} to show the training set accuracy, test set accuracy, and feature deviation per class along the training epochs for CIFAR-100, Tiny-ImageNet, and iNaturalist. 

\autoref{fig:fig2} shows the results on long-tailed CIFAR-100 ($\rho=100$). We see a similar trend as in \autoref{fig:2}: the training accuracy per class is almost $100\%$, while the test accuracy of minor classes is significantly low. We also see a clear trend of feature deviation: minor classes have larger deviation. Our CDT can compensate for the effect of feature deviation and leads to better test accuracy for minor classes.

\autoref{sup-fig:tiny} shows the results on long-tailed Tiny-ImageNet ($\rho=100$). We see a similar trend as in \autoref{fig:2}.

Since iNaturalist only has 3 validation images per class, there is a large variance in validation accuracy per class. We therefore re-split the data, keeping only classes that have at least $47$ training instances and move $22$ of them to the validation set. The resulting dataset has 1,462 classes and each class has at least $25$ training and $25$ validation instances. We then retrain the ConvNet classifier. \autoref{sup-fig:inat} shows the results. We see a similar trend as in \autoref{fig:2}.

\begin{figure*}[t]
	\centering
	\minipage{0.24\textwidth}
	\includegraphics[width=1.\linewidth]{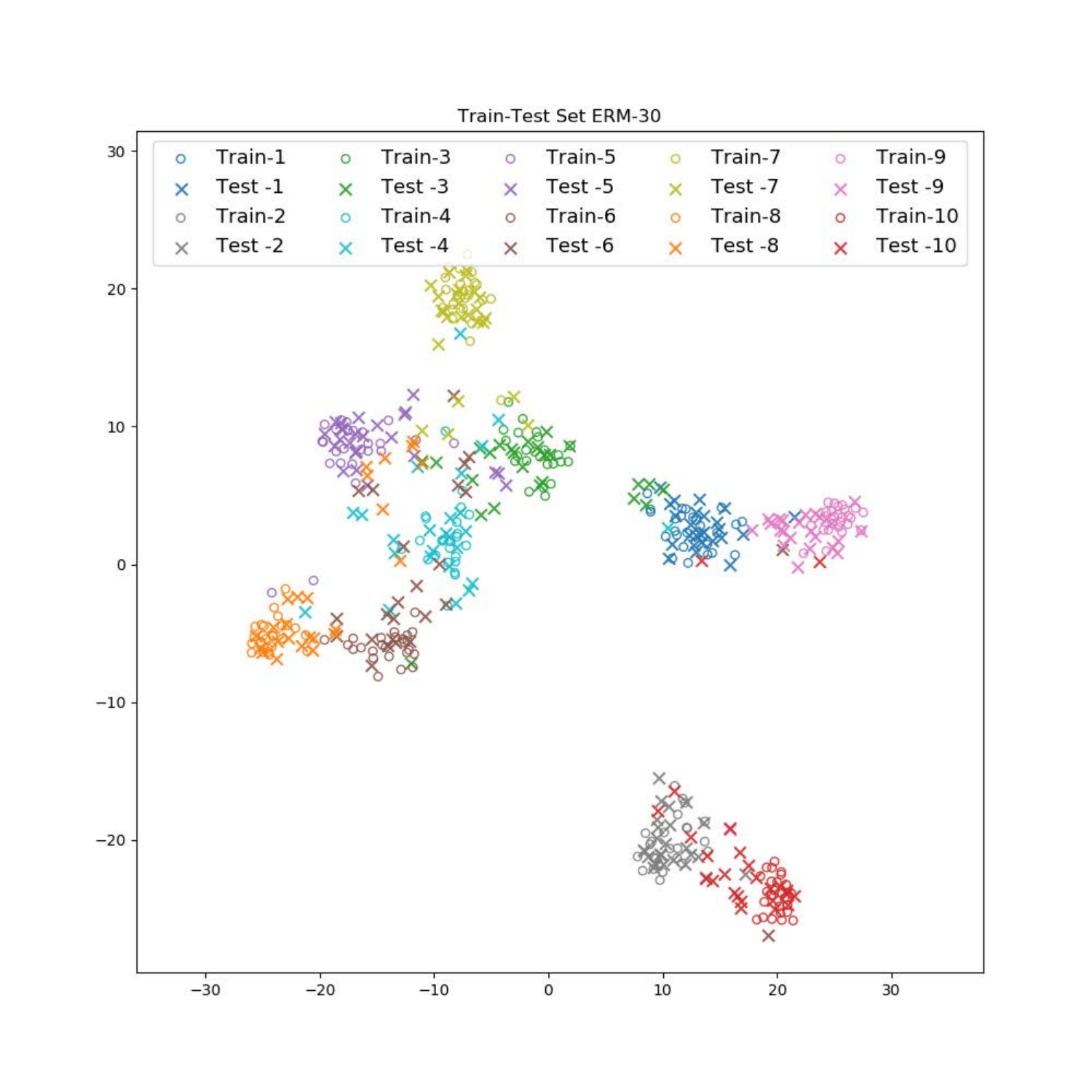}
	\centering
	\mbox{ERM}
	\endminipage
	\hfill
	\minipage{0.24\textwidth}
	\includegraphics[width=1.\linewidth]{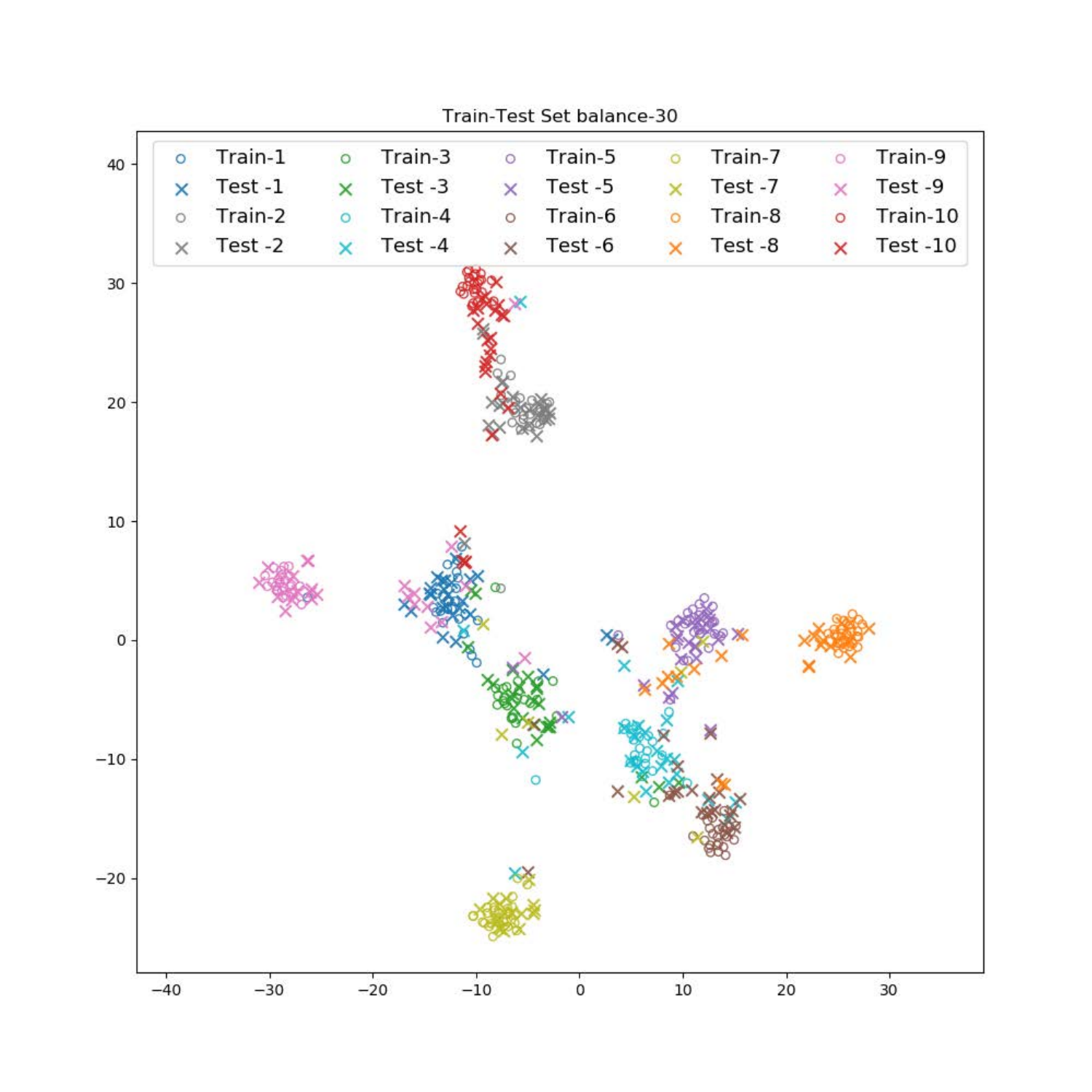}
	\centering
	\mbox{Re-sampling}
	\endminipage\hfill
	\minipage{0.24\textwidth}
	\includegraphics[width=1.\linewidth]{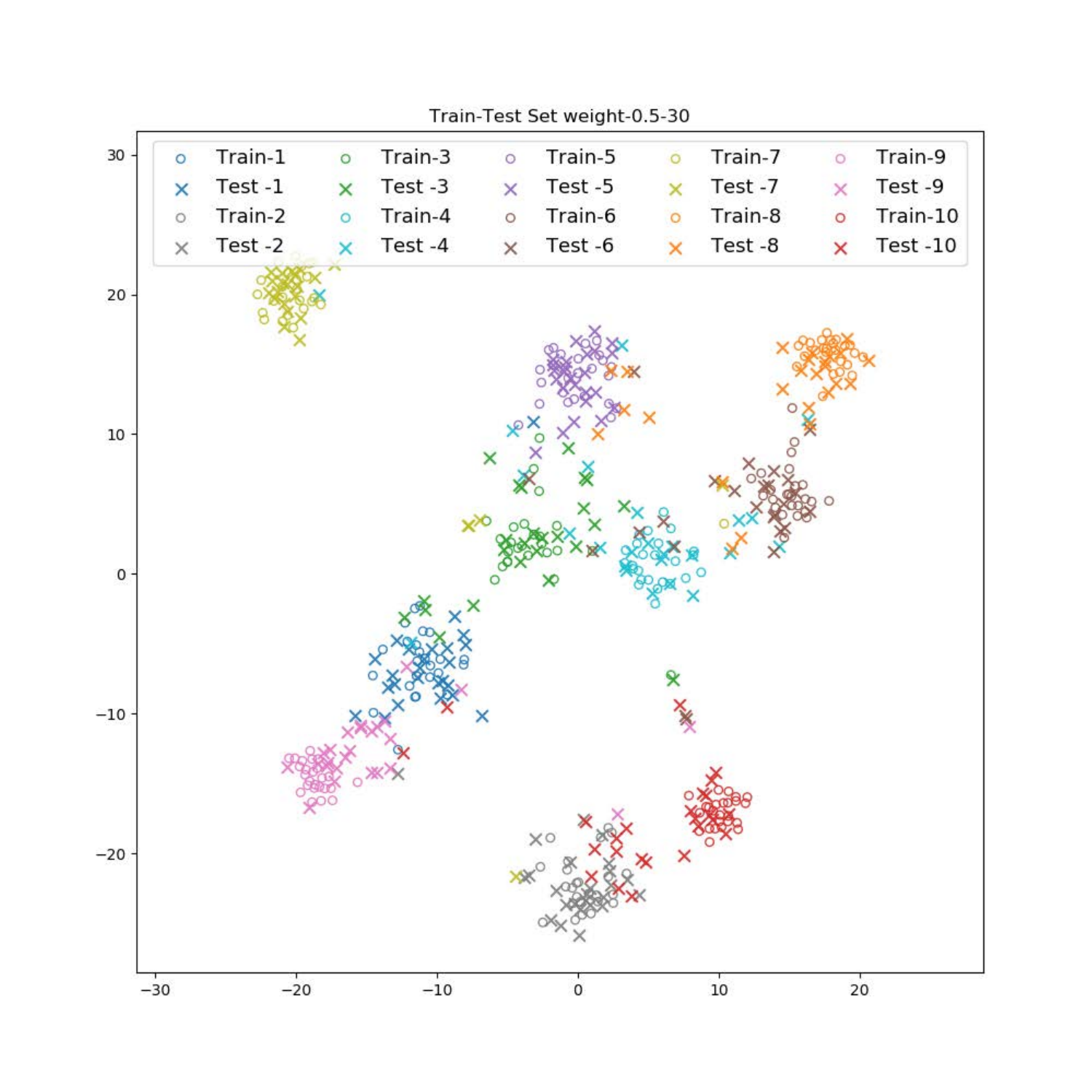}
	\centering
	\mbox{Re-weighting}
	\endminipage\hfill
	\minipage{0.24\textwidth}
	\includegraphics[width=1.\linewidth]{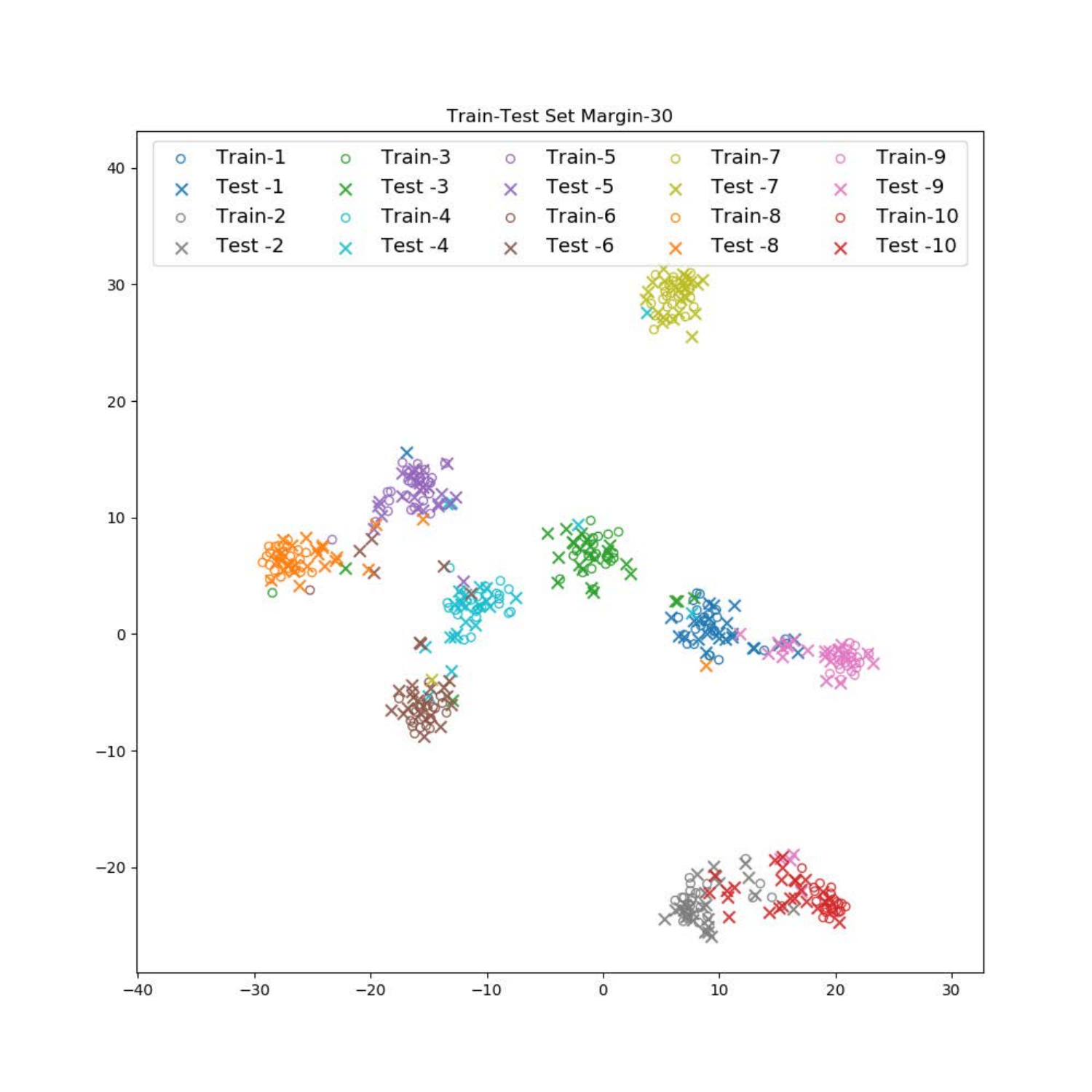}
	\centering
	\mbox{CDT}
	\endminipage
	\caption{t-SNE plots~\cite{maaten2009visualizing} of various methods on long-tailed CIFAR-10 ($\rho=100$). We sample 25 training (circle)/test (cross) instances per class.}
	\label{fig:tsne100}
\end{figure*}

\subsection{Feature visualization}
\label{suppl-ssec:tsne}

We provide additional feature visualization (t-SNE plots~\cite{maaten2009visualizing}) in~\autoref{fig:tsne100}. We sample 25 training (circle)/test (cross) instances per class and perform t-SNE. We see clear feature deviation of minor classes (\eg, see the red and magenta classes) for all the training objectives. Our CDT could separate different classes better than other approaches. Thus, even with feature deviation, the training and test instances of the same class are still closer than different classes.

\section{Discussions}
\label{sec:disc_disc}
\subsection{Imbalanced learning for deep learning and traditional machine learning}
In our pursuit to resolve imbalanced deep learning, we found that many existing works follow the strategies or principles of traditional machine learning to design algorithms. After a careful analysis we found that the reasons for poor minor-class accuracy are different: traditional machine learning suffers under-fitting; deep learning suffers over-fitting. It appears that such a difference is usually overlooked in imbalanced deep learning, and many algorithms are still designed to implicitly fit the minor class instances more.

We note that, the machine learning concept that small training data leads to over-fitting mainly applies to \emph{class-balanced learning (\ie, every class has small data)}. In imbalanced learning, traditional machine learning approaches like logistic regression with pre-defined features suffer under-fitting to minor classes (see \autoref{ssec:traditional_deep} and~\autoref{fig:linear}). In this case, re-weighting and re-sampling are effective: they make the model fit minor class instances more (see \autoref{fig:linear}). In imbalanced deep learning, however, we found that empirical risk minimization (ERM) already led to over-fitting to minor classes. While the seminal work by \cite{zhang2016understanding} on balanced learning may explain it, many imbalanced deep learning methods still apply re-weighting and over-sampling, which tend to further fit minor-class data. In other words, imbalanced deep learning is usually treated in the same way as imbalanced traditional machine learning, without noting that the reasons for poor accuracy are drastically different.

\subsection{CDT vs. re-weighting and post-calibration}
CDT is different from re-weighting. Let $(\vx, y)$ be a training instance and $l(\mW^\top f_{\vtheta}(\vx), y)$ be the loss function, re-weighting times $a_y$  to the loss (\ie, $a_y\times l(\mW^\top f_{\vtheta}(\vx), y)$), where minor classes have larger $a_y$. In contrast, CDT divides each $\vw_c$ in $\mW$ by $a_c$. As ConvNets can fit training data well (\ie, achieve low $l(\mW^\top f_{\vtheta}(\vx), y))$, training with or without re-weighting will not change the decision values much. In contrast, CDT forces $\vw_c^\top f_{\vtheta}(\vx)$ to be larger by $a_c$ to compensate for the decision value drop in testing due to feature deviation. 

CDT and post-calibration (\ie, $\tau$-normalization~\cite{kang2019decoupling}) adjust $\vw_c^\top f_{\vtheta}(\vx)$ differently. CDT divides those of {\em minor classes  more in training} (see \autoref{eq:margin} of the main paper), encouraging the classifier to learn to compensate for it. In contrast, post-calibration divides those of {\em major classes more in testing}. Specifically, $\tau$-normalization applies the following decision rule 
\begin{align}
	\hat{y} = & \argmax_{c} \cfrac{\vw_c^\top}{\|\vw_c\|_2^\tau} f_{\vtheta} (\vx), \label{eq:test_tau}
\end{align}
where $\tau\in[0,1]$ is a hyperparameter. We note that, major classes usually have larger $\|\vw_c\|_2$.

\subsection{The reason for feature deviation}
The image distributions of the training and test images are similar in the datasets we experimented with. The deviation is mainly caused by training ConvNets end-to-end with few minor-class images. These can be seen from \autoref{fig:3} (c): when the feature extractor is trained with balanced and sufficient data (ERM-UB) or pre-defined from other datasets (ERM-PD), there is no trend of feature deviation.

\section{Conclusion}
Classifiers trained with class-imbalanced data are known to perform poorly on minor classes. We perform a comprehensive analysis and find that the feature deviation phenomenon plays a crucial role in making a ConvNet classifier over-fit minor-class data. Such a phenomenon, however, is rarely observed in traditional machine learning approaches that use pre-defined features. To compensate for the effect of feature deviation --- which pushes test data of minor classes toward regions of low decision values --- we propose to incorporate class-dependent temperatures (CDT) in learning a ConvNet classifier. CDT is simple to implement and achieves promising results on benchmark datasets. We hope that our findings and analysis would inspire more advanced algorithms in class-imbalanced deep learning.
\label{sec:disc}


\ifCLASSOPTIONcompsoc
\section*{Acknowledgments}
\else
\section*{Acknowledgments}
\fi
This research is supported by National Key R\&D Program of China (2020AAA0109401), NSFC (61773198,61921006,62006112), NSFC-NRF Joint Research Project under Grant 61861146001, Collaborative Innovation Center of Novel Software Technology and Industrialization, NSF of Jiangsu Province (BK20200313), NSF IIS-2107077, NSF OAC-2118240, NSF OAC-2112606, and the OSU GI Development funds.
We are thankful for the generous support of computational resources by Ohio Supercomputer Center and AWS Cloud Credits for Research.

\ifCLASSOPTIONcaptionsoff
\newpage
\fi




{\small
\bibliographystyle{IEEEtran}
\bibliography{ref}
}

%

%

\end{document}